\documentclass[lettersize,journal]{IEEEtran}
\usepackage{amsmath,amsfonts}
\usepackage{algorithmic}
\usepackage{algorithm}
\usepackage{array}
\usepackage[caption=false,font=normalsize,labelfont=sf,textfont=sf]{subfig}
\usepackage{textcomp}
\usepackage{stfloats}
\usepackage{url}
\usepackage{verbatim}
\usepackage{graphicx}
\usepackage{cite}
\usepackage{tabularx}
\usepackage{booktabs}
\usepackage{makecell}
\usepackage{amssymb}
\usepackage{multirow}
\usepackage{nccmath}
\usepackage{pifont}
\usepackage[table]{xcolor}
\usepackage[colorlinks,linkcolor=blue,citecolor=blue,urlcolor=blue]{hyperref}
\begin{document}

\title{Hyperspectral Intrinsic Decomposition: Joint Recovery of Reflectance and Photometric Components for Non-Lambertian Scenes}

\author{
    Hao Ye,
    Zhan Shi,
    Chenglong Huang,
    Tao Lv,
    Mingjie Ji,
    Qiu Shen,~\IEEEmembership{Member,~IEEE,}
    and Xun Cao,~\IEEEmembership{Member,~IEEE}
    \thanks{Hao Ye, Zhan Shi, Chenglong Huang, Tao Lv, Mingjie Ji, Qiu Shen, and\\Xun Cao are with the School of Electronic Science and Engineering, Nanjing University, Nanjing 210023, China. E-mail: \{yehao, zhanshi, Chenglong-Huang, lvtao, jmj\}@smail.nju.edu.cn, \{shenqiu, caoxun\}@nju.edu.cn.}
    \thanks{Corresponding authors: Xun Cao.}
    \thanks{Code and dataset will be available at \url{https://github.com/CITE-NJU/DichroicFormer}.}
}


\markboth{IEEE Transactions on Pattern Analysis and Machine
Intelligence}{}

\maketitle

\begin{abstract}
Hyperspectral intrinsic decomposition (HID) aims to disentangle material-related spectral properties and photometric effects in hyperspectral images (HSIs), which is essential for understanding real-world imaging processes and benefits a variety of downstream applications. Most existing HID studies have been developed under Lambertian or near-Lambertian assumptions. The few prior non-Lambertian efforts rely on simplified specular assumptions insufficient to handle diverse real-world specularity, and typically require auxiliary inputs or recover only a subset of the coupled reflectance and photometric components, hindering complete and blind decomposition. In this paper, we revisit the dichromatic reflection model (DRM) and develop a unified inversion paradigm that reformulates the recovery of four coupled reflectance and photometric components as the estimation of two spectral--spatial target variables. Building on this reformulation, we propose a dual-scale decomposition scheme to handle non-Lambertian effects with distinct spatial characteristics. At the global scale, photometrically invariant descriptors serve as edge priors for high-fidelity intrinsic boundary preservation; at the local scale, specularity-guided attention directs refinement with emphasis on specularity-dominated regions, including those affected by clipping distortion. To facilitate future research, we establish CITE, the first public real-world HID dataset for non-Lambertian objects, and develop a Physically-faithful Intrinsic Set Generator (PISG) for controllable data synthesis. Extensive ablation studies and experiments on the CITE and additional HSIs demonstrate the effectiveness of our method and its robustness across diverse scenes.
\end{abstract}
\begin{IEEEkeywords}
Dichromatic reflection model, hyperspectral image processing,
intrinsic image decomposition, non-Lambertian surfaces,
specular reflection
\end{IEEEkeywords}

\section{Introduction}
\IEEEPARstart{G}{round-based} hyperspectral imaging has advanced rapidly with the miniaturization~\cite{ADIS, shi2023compact} and commercialization of spectrometers, enabling a wide range of real-world applications. Nevertheless, practical ground-based acquisition is often complicated by diverse scene geometry, artificial illumination, and specular reflections, which entangle material-related spectral properties with photometric effects, thereby hindering reliable fine-grained spectral analysis. Hyperspectral intrinsic decomposition (HID) provides a principled way to disentangle material-related reflectance and photometric effects in hyperspectral images (HSIs), yielding acquisition-invariant representations for downstream tasks such as agricultural monitoring~\cite{yao2024evaluation} and medical diagnosis~\cite{alotaibi2019decomposing}, as well as physically meaningful photometric cues for scene understanding~\cite{li2021multispectral, nam2014multispectral}.

Recent studies on HID under Lambertian assumptions~\cite{chen2017intrinsic, huang2018multispectral} have achieved notable progress, but their focus on diffuse reflectance decomposition leaves specular effects largely unaddressed. A few works have therefore extended single-image HID to non-Lambertian objects~\cite{huynh2008optimal, huynh2010solution, gu2013efficient, barron2014shape, krebs2020intrinsic}, typically using the dichromatic reflection model (DRM)~\cite{shafer1985using} to account for the additional specular component. However, these methods generally rely on handcrafted priors tailored to simplified manifestations of specular reflection. For example, Huynh and Robles-Kelly~\cite{huynh2010solution} treated specular reflection as an outlier within locally homogeneous diffuse regions; Gu~et~al.~\cite{gu2013efficient} neglected the diffuse component in strong specular regions; and Krebs~et~al.~\cite{krebs2020intrinsic} assumed specular reflection to be absent around reflectance edges.

\begin{figure*}[!t]
  \centering
  \includegraphics[width=0.9\textwidth, keepaspectratio]{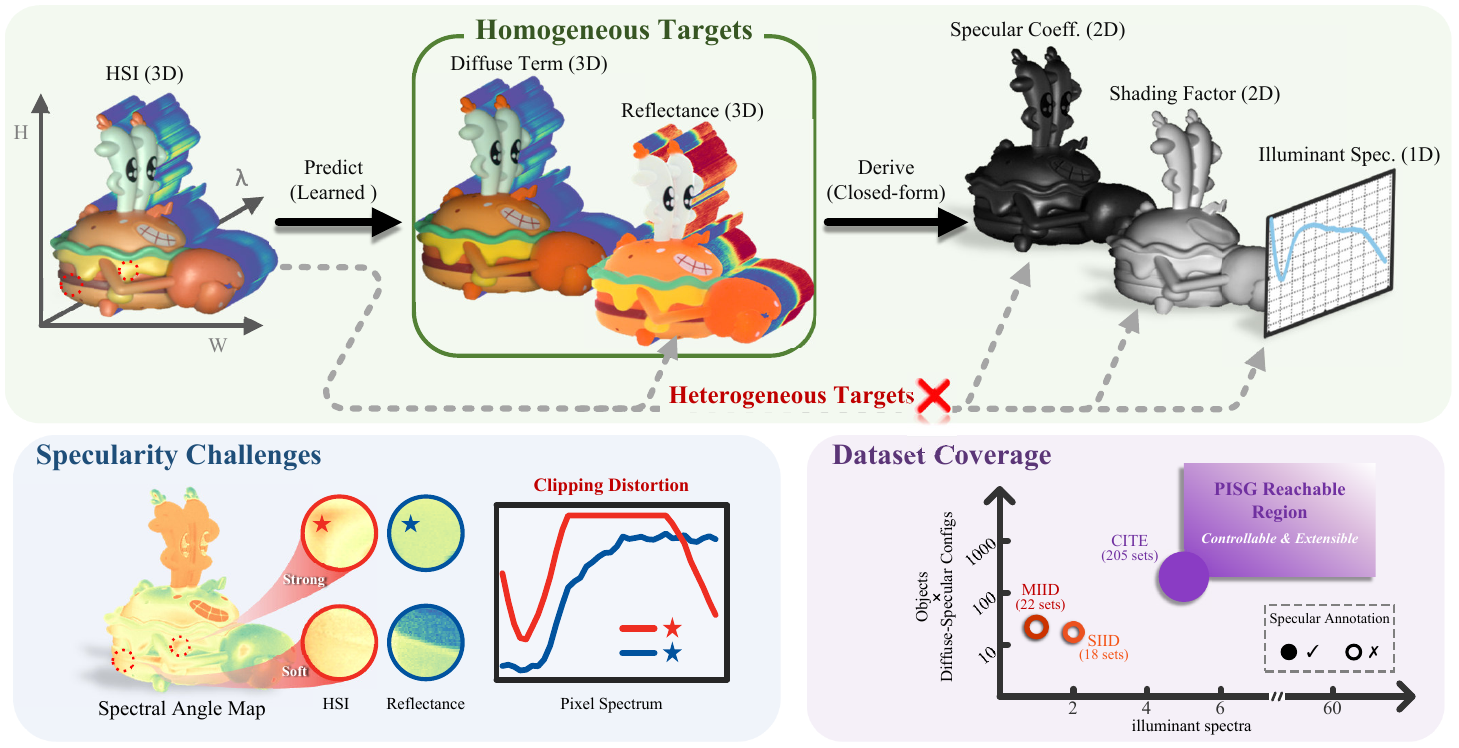}
    \caption{Hyperspectral intrinsic decomposition on non-Lambertian surfaces. \textbf{(Top)}~The dichromatic reflection model (DRM) couples four components with heterogeneous dimensions---3D reflectance, 2D shading factor, 2D specular coefficient, and 1D illuminant spectrum---complicating joint recovery (dashed gray path, \textcolor{red}{$\boldsymbol{\times}$}). We instead estimate two homogeneous spectral--spatial targets, namely the diffuse term and the reflectance, from which the illuminant spectrum and spatial photometric factors can be derived in closed form. \textbf{(Bottom-left)}~Strong specular reflections dominate local spectral signatures and may exceed the sensor dynamic range, causing clipping distortion (spectral plateau at \textcolor{red}{$\bigstar$}); soft specular reflections are spatially widespread and attenuate inter-material spectral differences. The per-pixel spectral angle map, computed using a uniform all-ones reference spectrum, shows that our decomposition improves intra-material spectral consistency and inter-material separation. \textbf{(Bottom-right)}~Existing real-world HID datasets (SIID~\cite{chen2017intrinsic}, MIID~\cite{huang2018multispectral}) lack specular-component annotations and are limited in object and illuminant diversity. Our CITE dataset provides component-level annotations for non-Lambertian objects with substantially broader coverage, and the Physically-faithful Intrinsic Set Generator (PISG) further enables controllable synthesis along both illuminant-spectrum and diffuse--specular-configuration axes (shaded region). The 3D spectral--spatial visualizations unfold the spectral dimension along the vertical axis with intensity-coded pseudo-color.}
  \label{fig:Introduction}
\end{figure*}

In practice, specular reflections exhibit diverse spatial and intensity characteristics~\cite{shi2016benchmark, fu2024towards}. As illustrated in Fig.~\ref{fig:Introduction}, soft specular reflections are low in intensity yet spatially widespread, attenuating material-related spectral differences and thereby blurring material boundaries in transitional regions. By contrast, sparse strong specular reflections can dominate local spectral signatures, creating apparent boundaries that are driven by specularity rather than material variation. In extreme cases, strong specular reflections may exceed the sensor dynamic range, causing clipping distortion~\cite{li2006estimating, nardecchia2021saturated} that breaks the consistency between the recorded spectra and the underlying dichromatic mixing process. The idealized assumptions adopted in previous methods are therefore insufficient to handle such complex and varied specular behaviors in real scenes.

An additional challenge arises from the structural heterogeneity of the DRM components: the reflectance is defined over both spatial and spectral dimensions, the shading and specular coefficients over the spatial dimension alone, and the illuminant spectrum over the spectral dimension alone. This heterogeneity complicates holistic decomposition, and combined with the aforementioned specular challenges, has led previous studies to focus on estimating only a subset of the coupled components, even though they remain physically coupled in the image formation process.

To address these challenges, we propose DichroicFormer, a single-image non-Lambertian HID framework built upon a unified inversion paradigm and a dual-scale decomposition scheme. Specifically, the inversion paradigm reformulates the recovery of four coupled reflectance and photometric components as the estimation of two spectral--spatial target variables, unifying the heterogeneous representation domains and providing a compact basis for complete joint recovery. Building on this reformulation, the dual-scale scheme addresses non-Lambertian effects with distinct spatial characteristics. At the global stage, the Invariant-Driven Module (IDM) exploits the Spectral-Gradient Ratio (SR) and Cross Spectral-Gradient Ratio (CSR), two descriptors with different levels of photometric invariance, as target-aligned edge priors for intrinsic boundary preservation. At the local stage, the Specularity-Guided Attention Module (SGAM) guides refinement with emphasis on specularity-dominated regions, including those affected by clipping distortion.

To support this study and future research, we establish CITE, the first public real-world HID dataset for non-Lambertian objects with component-level annotations, and develop a Physically-faithful Intrinsic Set Generator (PISG) for controllable synthesis under diverse illuminant spectra and diffuse--specular configurations. We further introduce scale-coupled evaluation metrics that account for the coupled scale relationships inherent in the DRM, enabling physically consistent assessment. The main contributions of this work are summarized as follows.
\begin{enumerate}
    \item We propose a unified inversion paradigm under the dichromatic reflection model that reformulates the recovery of four coupled reflectance and photometric components as the estimation of two spectral--spatial target variables, thereby unifying heterogeneous representation domains and providing a compact basis for complete joint recovery.
    \item We develop a dual-scale decomposition scheme to handle non-Lambertian effects with distinct spatial characteristics, where the IDM exploits descriptors with different levels of photometric invariance as edge priors for global intrinsic boundary preservation, and the SGAM guides local refinement with emphasis on specularity-dominated regions, including clipping-affected regions.
    \item We establish CITE, the first public real-world HID dataset for non-Lambertian objects with component-level annotations, and develop a Physically-faithful Intrinsic Set Generator (PISG) for controllable synthesis under diverse illuminant spectra and diffuse--specular configurations. Scale-coupled evaluation metrics are also introduced to enable physically consistent assessment under the coupled scale relationships inherent in the DRM. Together, these resources provide a systematic benchmark for non-Lambertian HID.
    \item Extensive experiments on CITE and additional hyperspectral images, together with comprehensive ablation studies, demonstrate the effectiveness of the proposed framework and its generalization capability across diverse non-Lambertian scenes.
\end{enumerate}

\section{Related Work}
\subsection{RGB Intrinsic Images Decomposition}
Intrinsic image decomposition in RGB images has been extensively studied since the seminal work of Barrow and Tenenbaum~\cite{barrow1978recovering}, which introduced the idea of representing distinct scene properties as separate intrinsic components. Early studies mainly relied on handcrafted priors to restrict the solution space. Among them, Retinex-based methods~\cite{land1971lightness} exploit color constancy to separate illumination from reflectance; a widely adopted complementary assumption~\cite{weiss2001deriving} further posits that gradual intensity variations are more likely caused by illumination variation, whereas abrupt changes often indicate reflectance boundaries. Other representative priors include reflectance sparsity~\cite{rother2011recovering, fu2019towards}, texture statistics~\cite{zhao2012closed, shen2013intrinsic, shen2008intrinsic}, and shading smoothness~\cite{li2014single, bi20151, sheng2018intrinsic}.

With the development of deep learning, RGB intrinsic decomposition has gradually shifted from prior-driven optimization to data-driven modeling. Supervised methods pioneered this transition by learning the implicit mapping between reflectance and shading directly from data~\cite{narihira2015direct}. Subsequent architectures based on Laplacian pyramids~\cite{liu2024enhancing, cheng2018intrinsic} and transformers~\cite{luo2023crefnet} have further improved decomposition fidelity. Beyond supervised learning, unsupervised frameworks~\cite{janner2017self, liu2020unsupervised} reduce reliance on paired annotations, while inverse-rendering-based approaches~\cite{liang2024eyeir} explicitly model the image formation process. Despite this progress, these methods are primarily designed for trichromatic RGB inputs and rely on color-space statistics or image cues tailored to low-dimensional representations. Compared with RGB images, HSIs provide substantially richer spectral information, but also introduce different decomposition variables and physical constraints. Consequently, existing designs for RGB intrinsic decomposition do not readily transfer to the hyperspectral domain, motivating the development of dedicated intrinsic decomposition methods for HSIs.

\subsection{Intrinsic Decomposition for Spectral Images}
Compared with its RGB counterpart, intrinsic decomposition for spectral images remains considerably less explored. The problem is also more challenging due to the higher-dimensional spectral observations and the need to jointly recover quantities defined over heterogeneous representation domains. Early studies mainly addressed this problem under Lambertian assumptions, exploiting the spectral structure to regularize the decomposition. Chen~et~al.~\cite{chen2017intrinsic} combined superpixel-based variable reduction with Retinex-inspired global sparsity and local constancy priors on reflectance, while Huang~et~al.~\cite{huang2018multispectral} further incorporated low-dimensional subspace constraints on both reflectance and shading spectra to improve problem tractability. These methods focus exclusively on diffuse reflectance decomposition, leaving the specular component unaddressed.

Extending spectral intrinsic decomposition to non-Lambertian objects requires explicitly modeling specular reflection, typically under the dichromatic reflection model (DRM)~\cite{shafer1985using}. Existing approaches primarily improve tractability by imposing simplified assumptions on specular behavior. Huynh and Robles-Kelly~\cite{huynh2008optimal, huynh2010solution} treated specular highlights as outliers within locally uniform reflectance regions, though this assumption becomes less reliable in textured areas. Gu~et~al.~\cite{gu2013efficient} neglected the diffuse component in strong specular regions, while Krebs~et~al.~\cite{krebs2017quadratic, krebs2020intrinsic} assumed specular reflection to be absent near material boundaries and reformulated DRM inversion into decoupled optimization steps. Rahman and Robles-Kelly~\cite{rahman2013optimisation, rahman2017estimating} introduced micro-geometric surface-reflection parameters for decomposition, but the smoothness constraints imposed on the microfacet slope limit performance in regions with high-frequency surface detail. 

Despite the progress made by these methods, the statistical regularities or handcrafted priors they rely on oversimplify specular behavior, making them difficult to adapt to real-world scenes where soft and strong specularities coexist. Moreover, the reflectance and photometric components in the DRM span heterogeneous representation domains, including spectral, spatial, and spectral--spatial quantities, which further increases the difficulty of joint inversion and has led previous studies to focus on estimating only a subset of the coupled components while relying on external estimation for the remainder.

A separate line of work further reduces ambiguity by introducing auxiliary measurements such as depth cues~\cite{mehami2022multi}, optical flow~\cite{bergamasco2017spectral}, multi-view observations~\cite{sinha2024spectralgaussians}, or specially designed illumination~\cite{guo2021multispectral, guo2022multispectral}, but these require multiple observations, controlled acquisition setups, or dedicated hardware, placing them outside the scope of this work.

Therefore, single-image non-Lambertian HID that achieves complete and blind recovery of all coupled components under complex specular effects remains insufficiently addressed.

\subsection{Datasets and Evaluation}
Benchmarking spectral intrinsic decomposition requires both annotated datasets and appropriate evaluation protocols, yet both remain underdeveloped for non-Lambertian settings. On the dataset side, several synthetic datasets with ground-truth scene attributes have been introduced in multispectral photometric stereo research~\cite{guo2022multispectral, lv2023non, zhou2020shape}. However, these datasets rely on specialized multi-illumination setups that are incompatible with the single-image setting; moreover, the domain gap between purely rendered data and real-world observations can further limit the practical effectiveness of data-driven methods. For real-world data, Chen~et~al.~\cite{chen2017intrinsic} constructed one of the earliest public datasets, SIID, acquiring annotations for nine objects under two illuminants of distinct color temperatures. Although the paper reports specularity, the publicly released annotations do not include an explicit ground-truth specular-reflection component. Huang~et~al.~\cite{huang2018multispectral} presented the MIID dataset, capturing 22 objects with a snapshot spectrometer under warm incandescent illumination, likewise providing only Lambertian annotations. Collectively, these real-world datasets remain limited in object diversity, illumination coverage, and annotation completeness: none provides ground-truth specular components, with at most two illumination conditions and fewer than thirty annotated sets per dataset.

On the evaluation side, commonly used metrics, such as scale-invariant MSE (si-MSE)~\cite{krebs2020intrinsic} and local MSE (LMSE)~\cite{huang2018multispectral, krebs2020intrinsic_Elsevier}, typically allow independent scale alignment for individual predicted components before error computation. While this practice compensates for the inherent scale ambiguity in the decomposition, it effectively treats components as independent regression targets, neglecting the physical coupling imposed by the image formation model. As a result, relative scale violations between components may go undetected even when individual component-wise errors appear acceptable. Such violations directly affect image-formation consistency and distort inter-component intensity ratios, thereby reducing the fidelity of evaluation under non-Lambertian settings.

\begin{figure*}[!t]
  \centering
  \includegraphics[width=0.9\textwidth,keepaspectratio]{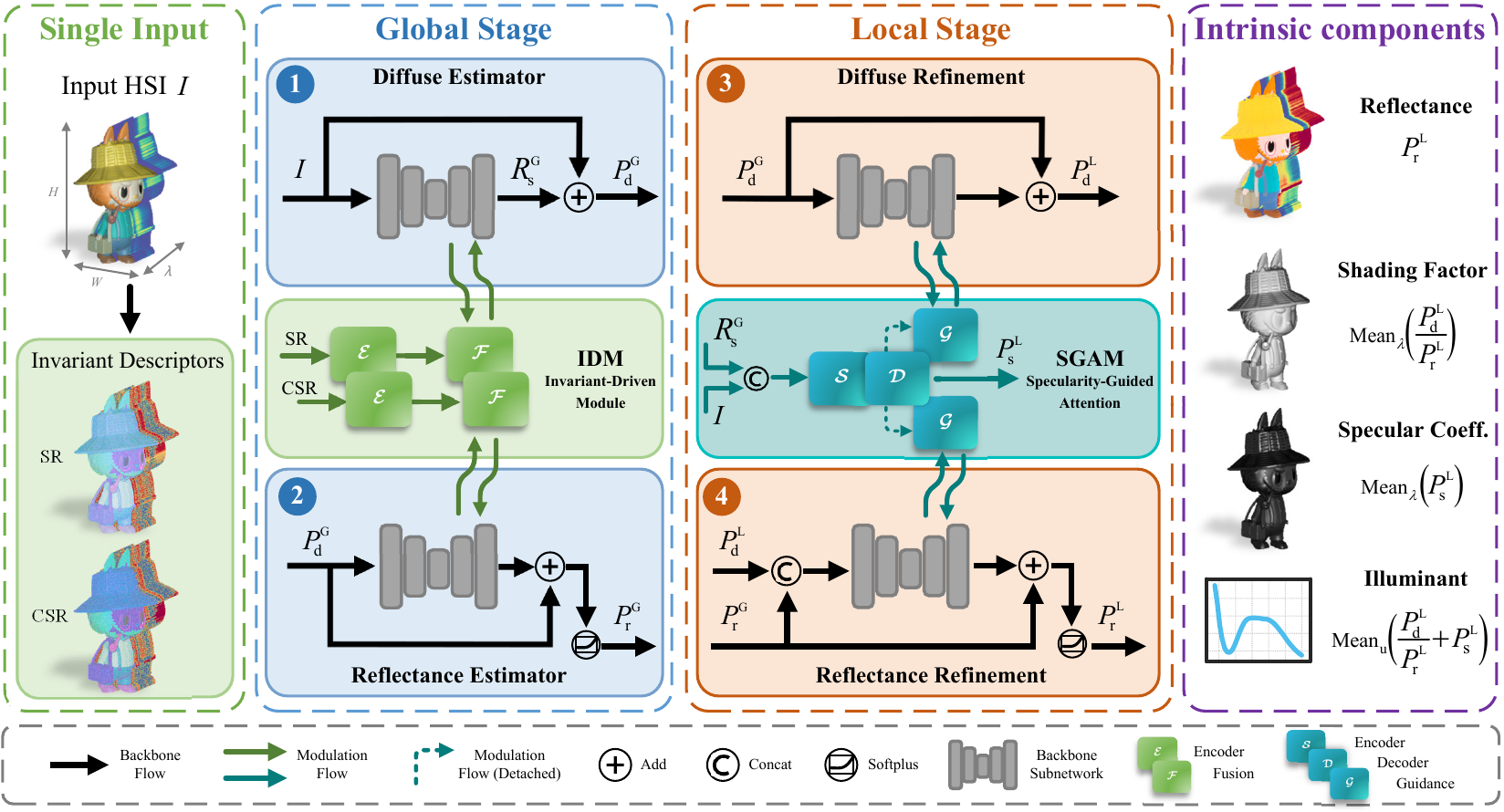}
    \caption{Overview of the proposed dual-scale decomposition framework. \textbf{(Single Input)}~The input HSI $I$ and two photometrically invariant descriptors (SR, CSR) computed from $I$ under the DRM. \textbf{(Global Stage)}~Subnetworks~\textcolor[RGB]{46,117,181}{\ding{202}\ding{203}} progressively estimate the diffuse term $P_\mathrm{d}^\mathrm{G}$ and reflectance $P_\mathrm{r}^\mathrm{G}$ following the decomposition order of the inversion paradigm. The Invariant-Driven Module (IDM) encodes SR and CSR through dedicated encoding and fusion paths ($\mathcal{E}, \mathcal{F}$) and fuses descriptor-based boundary priors with multi-level backbone features at each decoder level, with each descriptor routed to the subnetwork matching its level of photometric invariance. \textbf{(Local Stage)}~Subnetworks~\textcolor[RGB]{197, 90, 17}{\ding{204}\ding{205}} refine the global-stage estimates into $P_\mathrm{d}^\mathrm{L}$ and $P_\mathrm{r}^\mathrm{L}$. The Specularity-Guided Attention Module (SGAM) takes $R_\mathrm{s}^\mathrm{G}$ and $I$ as input. Its U-shaped specular subnetwork ($\mathcal{S}, \mathcal{D}$) produces specularity-aware features that, after detachment, modulate the encoder-to-decoder feature flow through $\mathcal{G}$. Meanwhile, the subnetwork provides the final specular prediction $P_\mathrm{s}^\mathrm{L}$ through an independently supervised pathway, decoupling the specular objective from the diffuse estimate under clipping-affected observations. \textbf{(Intrinsic Components)}~Reflectance corresponds to $P_\mathrm{r}^\mathrm{L}$; the shading factor, specular coefficient, and illuminant spectrum are derived in closed form from $\{P_\mathrm{d}^\mathrm{L}, P_\mathrm{r}^\mathrm{L}, P_\mathrm{s}^\mathrm{L}\}$.}
  \label{fig:Method_Pipeline}
\end{figure*}

\section{Problem Formulation and Inversion Paradigm}
\label{sec:INVERSION PARADIGM}
In this work, we assume the presence of either a single light source or multiple light sources sharing the same normalized spectral distribution, such that illumination varies across pixels only in intensity. We adopt the dichromatic reflection model (DRM) \cite{shafer1985using} to describe the image formation of non-Lambertian HSIs. Under the neutral interface reflection (NIR) assumption, the specular reflection preserves the spectral profile of the illuminant. The observed hyperspectral signal $I(u,\lambda)$ is then expressed as
\begin{equation}
\label{eq:drm}
I(u,\lambda) = g(u)\,L(\lambda)\,S(u,\lambda) + k(u)\,L(\lambda),
\end{equation}
where $u$ and $\lambda$ denote the pixel location and wavelength, respectively. $L(\lambda)$ denotes the illuminant spectrum, and $S(u,\lambda)$ is the spectral reflectance, i.e., the body/diffuse reflectance term in the DRM. The wavelength-independent scalars $g(u)$ and $k(u)$ are the shading factor and specular coefficient, respectively, which characterize the spatially varying contributions of the diffuse and specular reflections and capture the aggregate effects of surface geometry, viewing direction, and material glossiness.

A key difficulty of non-Lambertian HID is that the quantities involved in the DRM span heterogeneous representational domains: $S$ is defined over both spatial and spectral dimensions, $L$ over the spectral dimension alone, and $g$, $k$ over the spatial dimension alone. Directly recovering all four in parallel would therefore require handling structurally mismatched outputs, which may complicate model design and joint recovery. To address this issue, we reformulate the inversion around two spectral--spatial target variables, defined as
\begin{equation}
\label{eq:reparam}
\begin{aligned}
P_\mathrm{r}(u,\lambda) &= S(u,\lambda),\\
P_\mathrm{d}(u,\lambda) &= g(u)\,L(\lambda)\,P_\mathrm{r}(u,\lambda),
\end{aligned}
\end{equation}
where $P_\mathrm{r}$ represents the reflectance and $P_\mathrm{d}$ the diffuse term, with the latter aggregating reflectance under shading and illumination modulation. 

Substituting Eq.~\eqref{eq:reparam} into Eq.~\eqref{eq:drm} yields the additive form
\begin{equation}
\label{eq:formation}
I = P_\mathrm{d} + P_\mathrm{s},
\end{equation}
where $P_\mathrm{s}(u,\lambda) = k(u)\,L(\lambda)$ denotes the specular term, and arithmetic between these arrays is understood element-wise throughout this section. Since $P_\mathrm{s} = I - P_\mathrm{d}$ follows directly, the inversion task reduces to estimating $\{P_\mathrm{r},\, P_\mathrm{d}\}$, from which all remaining quantities can be recovered in closed form.

Following common practice in spectral imaging, the illuminant spectrum is recovered by spatial averaging with peak normalization:
\begin{equation}
\label{eq:illumination}
L = \frac{1}{\sigma M}\sum_{u}
\left(\frac{P_\mathrm{d}}{P_\mathrm{r}} + P_\mathrm{s}\right),
\sigma = \left\|\frac{1}{M}\sum_{u}
\left(\frac{P_\mathrm{d}}{P_\mathrm{r}} + P_\mathrm{s}\right)
\right\|_\infty,
\end{equation}
so that $\|L\|_\infty=1$. The shading-factor and specular-coefficient maps are then obtained by spectral averaging:
\begin{equation}
\label{eq:gk}
g = \frac{1}{N}\sum_{\lambda}
\frac{P_\mathrm{d}}{P_\mathrm{r}}, \qquad
k = \frac{1}{N}\sum_{\lambda} P_\mathrm{s},
\end{equation}
where $M$ and $N$ denote the total number of spatial pixels and spectral bands, respectively. This avoids per-wavelength division by $L$, which would otherwise be numerically unstable at wavelengths where $L(\lambda)$ is small, while preserving the relative magnitudes of $g$ and $k$. The inversion paradigm thus reduces the recovery of four heterogeneous components to the estimation of two spectral--spatial target variables, with all other quantities derived in closed form.

\section{Dual-Scale Scheme}
Specular reflections in real scenes exhibit distinct spatial and intensity characteristics. Soft specular reflections are typically low in intensity and spatially widespread, and tend to weaken material-related spectral discrimination over broad regions. By contrast, strong specular reflections are sparse and spatially concentrated, and may dominate the local spectral response. In extreme cases, they further induce clipping distortion when the sensor dynamic range is exceeded. Motivated by these distinct behaviors, we develop a dual-scale decomposition scheme for progressive non-Lambertian HID, consisting of a global stage and a local stage, as illustrated in Fig.~\ref{fig:Method_Pipeline}. The global stage focuses on high-fidelity decomposition with emphasis on preserving intrinsic boundaries, whereas the local stage further refines the decomposition, particularly in specularity-dominated regions. Both stages are built upon the same backbone architecture but independently parameterized, while scale-specific modules are introduced within the decoder pathway to incorporate stage-specific priors. Specifically, the global stage employs the Invariant-Driven Module (IDM, Sec.~\ref{sec:IDM}), and the local stage employs the Specularity-Guided Attention Module (SGAM, Sec.~\ref{sec:SGAM}). For notational clarity, superscripts $\mathrm{G}$ and $\mathrm{L}$ are used to distinguish outputs of the global and local stages, respectively.


\subsection{Backbone Architecture}
The backbone design follows the decomposition order implied by the proposed inversion paradigm: removing the additive specular term from the observation isolates the diffuse term $P_\mathrm{d}$, on the basis of which $P_\mathrm{r}$ can be more reliably estimated without specular interference. Accordingly, each stage employs a cascaded serial backbone composed of two U-shaped subnetworks, with the diffuse term and reflectance serving as progressive decomposition targets. Both subnetworks adopt a residual learning structure: the residual branch of the first subnetwork predicts the specular residual with respect to its input, which is added back to form a diffuse estimate approximating $P_{\mathrm{d}}$; the second subnetwork subsequently estimates the reflectance from this diffuse estimate via the same mechanism, with a positivity-preserving activation to avoid singularities in the subsequent division by $P_{\mathrm{r}}$ (cf. Eqs.~\eqref{eq:illumination} and~\eqref{eq:gk}).

In the global stage, the first subnetwork maps the input HSI $I$ to a diffuse estimate $P_{\mathrm{d}}^{\mathrm{G}} = I + R_{\mathrm{s}}^{\mathrm{G}}$; the second subnetwork then produces the reflectance estimate $P_{\mathrm{r}}^{\mathrm{G}}$ from $P_{\mathrm{d}}^{\mathrm{G}}$. In the local stage, the first subnetwork operates on \(P_\mathrm{d}^{\mathrm{G}}\) to yield the refined diffuse output \(P_\mathrm{d}^{\mathrm{L}} = P_\mathrm{d}^{\mathrm{G}} + R_\mathrm{s}^{\mathrm{L}}\). The second subnetwork takes the concatenation of \(P_\mathrm{d}^{\mathrm{L}}\) and \(P_\mathrm{r}^{\mathrm{G}}\) as input, leveraging the first-stage reflectance as an additional global prior, and outputs the final reflectance estimate \(P_\mathrm{r}^{\mathrm{L}}\). Based on the inversion relations established in Sec.~\ref{sec:INVERSION PARADIGM}, the refined outputs $P_{\mathrm{d}}^{\mathrm{L}}$ and $P_{\mathrm{r}}^{\mathrm{L}}$ constitute the two spectral--spatial targets of the inversion paradigm. In practice, the analytic relation $P_{\mathrm{s}} = I - P_{\mathrm{d}}$ is replaced by a learned specular prediction $P_{\mathrm{s}}^{\mathrm{L}}$ from SGAM (Sec.~\ref{sec:SGAM}), which decouples the specular objective from the diffuse estimate. The photometric quantities are then derived from $\{P_{\mathrm{d}}^{\mathrm{L}},\, P_{\mathrm{r}}^{\mathrm{L}},\, P_{\mathrm{s}}^{\mathrm{L}}\}$ via Eqs.~\eqref{eq:illumination} and~\eqref{eq:gk}, completing the joint recovery under the proposed paradigm.

Each U-shaped subnetwork adopts an encoder--bottleneck--decoder architecture. At each decoder level, the upsampled features are fused with the corresponding encoder features through the scale-specific module (IDM or SGAM; Secs.~\ref{sec:IDM} and~\ref{sec:SGAM}), and then refined by a transformer block. Placing the scale-specific module at the decoder side allows stage-specific priors to directly compensate for the information loss caused by downsampling during hierarchical reconstruction. Within each transformer block, Spectral-wise Multi-head Self-Attention (S-MSA) models long-range spectral dependencies, and the Gated-DConv Feed-Forward Network (GDFN) provides adaptive nonlinear feature transformation~\cite{cai2022mst++, zamir2022restormer}.

\subsection{Global-Scale Invariant-Driven Module}
\label{sec:IDM}
The global stage targets high-fidelity decomposition with faithful boundary preservation. To this end, following the cascaded decomposition targets established in the backbone, we derive two descriptors with varying levels of photometric invariance from the DRM, namely the Spectral-Gradient Ratio~($\mathrm{SR}$) and the Cross Spectral-Gradient Ratio~($\mathrm{CSR}$). Computed directly from the input HSI, these descriptors are incorporated into the proposed Invariant-Driven Module~(IDM) as edge-aware priors. Previous studies have explored invariant representations based on spectral gradients under known illumination conditions~\cite{montoliu2005illumination, ibrahim2010spectral}. In contrast, we extend this line of work to unknown illumination settings.

Given two adjacent spatial locations $u_1$ and $u_2$, together with an adjacent spectral-channel pair $(\lambda_1,\lambda_2)$, we first define the spectral gradient at location $u$ as
\begin{equation}
\label{eq:grad_def}
\operatorname{Grad}(u,\lambda_1,\lambda_2)=I(u,\lambda_1)-I(u,\lambda_2).
\end{equation}

Based on this quantity, the Spectral-Gradient Ratio is defined as
\begin{equation}
\label{eq:sr_def_1}
\mathrm{SR}(u_1,u_2,\lambda_1,\lambda_2) = 
\frac{\operatorname{Grad}(u_1,\lambda_1,\lambda_2)}
{\operatorname{Grad}(u_2,\lambda_1,\lambda_2)}.
\end{equation}

Substituting the DRM formulation in Eq.~\eqref{eq:drm} into Eq.~\eqref{eq:grad_def} gives
\begin{equation}
\label{eq:grad_expand}
\begin{aligned}
\operatorname{Grad}(u, \lambda_1, \lambda_2)
=\;& L(\lambda_1)\bigl[g(u)\,S(u,\lambda_1) + k(u)\bigr] \\
& - L(\lambda_2)\bigl[g(u)\,S(u,\lambda_2) + k(u)\bigr].
\end{aligned}
\end{equation}

To characterize inter-channel illuminant variation, we introduce an intermediate channel $\lambda_{1.5}$ midway between $\lambda_1$ and $\lambda_2$, and define
\begin{equation}
\label{eq:rho_def}
\rho_1 = \frac{L(\lambda_1)}{L(\lambda_{1.5})}, \qquad
\rho_2 = \frac{L(\lambda_2)}{L(\lambda_{1.5})}.
\end{equation}

Then, Eq.~\eqref{eq:grad_expand} can be rearranged as
\begin{equation}
\label{eq:grad_exact}
\begin{split}
\operatorname{Grad}(u, \lambda_1, \lambda_2) = & L(\lambda_{1.5}) g(u) [S(u,\lambda_1) - S(u,\lambda_2)] \\
& + \varepsilon(u, \lambda_1, \lambda_2),
\end{split}
\end{equation}
where the residual term $\varepsilon$ collects all contributions that depend on inter-channel illuminant variation:
\begin{equation}
\label{eq:epsilon_def}
\begin{aligned}
\varepsilon(u, \lambda_1, \lambda_2) = & L(\lambda_{1.5}) \Bigl[ g(u) \bigl( (\rho_1 - 1)S(u,\lambda_1) \\
& + (1 - \rho_2)S(u,\lambda_2) \bigr) + k(u)(\rho_1 - \rho_2) \Bigr].
\end{aligned}
\end{equation}

For typical visible-range HSIs with spectral sampling interval on the order of a few nanometers, the half-channel spacing between \(\lambda_{1.5}\) and either \(\lambda_1\) or \(\lambda_2\) is sufficiently small that the illuminant spectrum can be regarded as approximately constant over this interval, yielding $\rho_1 \approx \rho_2 \approx 1$, so that $\varepsilon \approx 0$. The spectral gradient thereby reduces to
\begin{equation}
\label{eq:grad_approx}
\operatorname{Grad}(u, \lambda_1, \lambda_2) 
\approx L(\lambda_{1.5})\, g(u) 
\bigl[S(u,\lambda_1) - S(u,\lambda_2)\bigr],
\end{equation}
from which the $\mathrm{SR}$ simplifies to
\begin{equation}
\label{eq:SR_simplified}
\mathrm{SR}(u_1, u_2, \lambda_1, \lambda_2) \approx \frac{g(u_1)\bigl[S(u_1,\lambda_1) - S(u_1,\lambda_2)\bigr]}{g(u_2)\bigl[S(u_2,\lambda_1) - S(u_2,\lambda_2)\bigr]}.
\end{equation}

This formulation shows that $\mathrm{SR}$ suppresses the influence of the illuminant spectrum and the additive specular term, while preserving the shading-modulated reflectance differences between adjacent pixels. It therefore serves as an effective boundary cue for the global decomposition stage. Furthermore, by introducing a third spectral channel \(\lambda_3\) adjacent to \(\lambda_2\), we define the Cross Spectral-Gradient Ratio as
\begin{equation}
\label{eq:CSR_1}
\begin{aligned}
\mathrm{CSR} &= \frac{\mathrm{SR}(u_1, u_2, \lambda_1, \lambda_2)}
{\mathrm{SR}(u_1, u_2, \lambda_2, \lambda_3)}
\\[4pt] &\approx \frac{\bigl(S(u_1,\lambda_1) - S(u_1,\lambda_2)\bigr)
\bigl(S(u_2,\lambda_2) - S(u_2,\lambda_3)\bigr)}
{\bigl(S(u_1,\lambda_2) - S(u_1,\lambda_3)\bigr)
\bigl(S(u_2,\lambda_1) - S(u_2,\lambda_2)\bigr)}.
\end{aligned}
\end{equation}

The $\mathrm{CSR}$ depends solely on the reflectance $S$ and is invariant to all photometric parameters in the DRM, thereby providing a more purely reflectance-driven description of spatial variation than $\mathrm{SR}$.

\begin{figure}[!t]
  \centering
  \includegraphics[width=0.85\columnwidth]{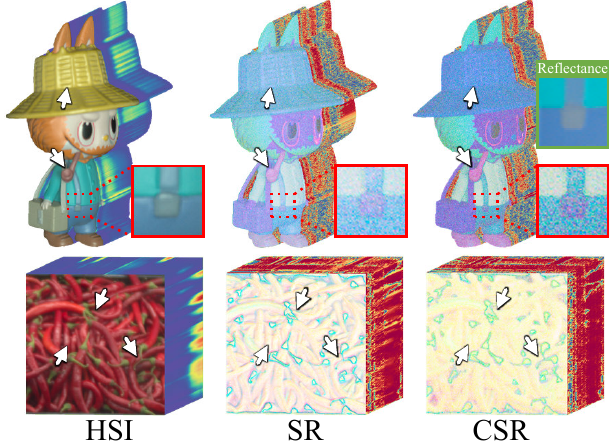} 
    \caption{HSI and the corresponding $\mathrm{SR}$ and $\mathrm{CSR}$ visualizations for two representative scenes. Arrows indicate selected local details for comparison. The \textcolor[RGB]{255, 0, 0}{red} box highlights a region where soft specularity attenuates inter-material spectral differences in the original HSI; the zoomed insets show that both $\mathrm{SR}$ and $\mathrm{CSR}$ reveal these contrasts more clearly, as confirmed by comparison with the ground-truth reflectance (\textcolor[RGB]{112, 173, 71}{green} border). Compared with the original HSI, the $\mathrm{SR}$ maps suppress the influence of the illuminant spectrum and specular reflection while retaining shading-modulated structure, whereas the $\mathrm{CSR}$ maps further suppress the residual shading dependence, thereby isolating reflectance-driven spatial boundaries.}
  \label{fig:SR_CSR}
\end{figure}

As illustrated in Fig.~\ref{fig:SR_CSR}, the $\mathrm{SR}$ and $\mathrm{CSR}$ computed for both scenes exhibit boundary structures associated with intrinsic material differences, indicating that they can effectively suppress the influence of specular reflection. This effect is particularly evident around the belt buckle in Fig.~\ref{fig:SR_CSR}(a): comparison with the zoomed reflectance details shows that spectral contrasts attenuated in the original HSI are more clearly revealed in both \(\mathrm{SR}\) and \(\mathrm{CSR}\). Meanwhile, because the $\mathrm{SR}$ remains dependent on the shading factor $g$, it additionally preserves texture cues associated with scene geometry, whereas such effects are further suppressed in the $\mathrm{CSR}$.

These two descriptors, with their different levels of photometric invariance, naturally align with the two decomposition targets of the backbone: $\mathrm{SR}$ retains both reflectance and shading structures and is therefore paired with the first subnetwork targeting the diffuse term $P_{\mathrm{d}}$, while $\mathrm{CSR}$, being purely reflectance-driven, is paired with the second subnetwork targeting the reflectance $P_{\mathrm{r}}$. Based on these observations, we introduce the Invariant-Driven Module~(IDM) to inject descriptor-based boundary priors at the decoder side of the global stage. As depicted in Fig.~\ref{fig:Method_Pipeline}, each input descriptor is first encoded through a lightweight layer for denoising and feature embedding, and then progressively fused with the corresponding multi-level encoder and decoder features, thereby driving the propagation of intrinsic boundary cues during hierarchical reconstruction. Downsampling within the module ensures spatial alignment with the multi-scale backbone features.

The reliability of these descriptors, however, depends on the approximation that $\varepsilon$ remains sufficiently small. As indicated by Eq.~\eqref{eq:epsilon_def}, although the inter-channel illuminant deviations are small under high spectral resolution, the residual can nonetheless become non-negligible in regions dominated by strong specularity, compromising the reliability of the descriptor-based priors; in extreme cases, the specular intensity may exceed the sensor dynamic range, further introducing clipping distortion and irreversible spectral information loss. Dedicated refinement is therefore still required for these sparse but locally concentrated regions, which motivates the SGAM introduced in the following subsection.

\subsection{Local-Scale Specularity-Guided Attention Module}
\label{sec:SGAM}
To address the limitations identified above, the local stage introduces the Specularity-Guided Attention Module~(SGAM), which leverages specular features from a dedicated subnetwork to guide backbone refinement with particular emphasis on specularity-dominated regions, while also providing a direct prediction pathway for the final specular term.

As depicted in Fig.~\ref{fig:Method_Pipeline}, the SGAM contains a compact U-shaped specular subnetwork that takes the global-stage specular residual $R_\mathrm{s}^{\mathrm{G}}$ together with the original HSI $I$ as input. The former provides a coarse spatial prior for specular localization, while the latter supplies the original content prior to any decomposition. Supervised by the ground-truth specular term, the subnetwork produces multi-level specularity-aware features that modulate the encoder-to-decoder feature flow within the backbone, providing additional local guidance during hierarchical reconstruction and enhancing the network's sensitivity to regions dominated by strong specular reflection. The specularity-aware features are detached before being injected into the backbone, ensuring that the specular subnetwork is optimized exclusively through its own supervision.

Meanwhile, the output of the subnetwork's mapping layer is taken as the final specular prediction $P_\mathrm{s}^{\mathrm{L}}$ of the local stage. In principle, following the analytic relation established in Sec.~\ref{sec:INVERSION PARADIGM}, $P_{\mathrm{s}}$ can also be obtained as $I - P_{\mathrm{d}}^{\mathrm{L}}$. However, this analytic pathway ties the final specular output directly to the diffuse estimate. In clipping-affected regions, where the recorded intensity is truncated below the true sum of the diffuse and specular terms, the diffuse and analytic specular objectives impose conflicting requirements on the same $P_\mathrm{d}^{\mathrm{L}}$. By routing the specular prediction through an independently supervised subnetwork, the SGAM decouples the specular objective from the diffuse pathway, allowing each to be optimized without interference and mitigating the analytic-pathway conflict under clipping-affected observations.

\subsection{Loss Function}
Both stages are trained with standard MSE losses. In each stage, supervision is imposed on the two direct network outputs, namely the diffuse term $P_{\mathrm{d}}$ and the reflectance $P_{\mathrm{r}}$, as well as on the ratio $P_\mathrm{d}/P_\mathrm{r}$, which represents the illumination-modulated shading. By jointly supervising these coupled quantities, the training objective enforces their mutual consistency, facilitating more balanced recovery of reflectance and the photometric components. In the local stage, an additional loss term is applied to the SGAM specular prediction $P_{\mathrm{s}}^{\mathrm{L}}$, which not only constrains the final specular output but also regularizes the learning of the multi-level specularity-aware features within the SGAM. The total training objective is a weighted sum of the global-stage and local-stage losses; detailed formulations and hyperparameter settings are provided in the supplementary material.

\begin{figure*}[!t]
  \centering
  \includegraphics[width=0.9\textwidth,keepaspectratio]{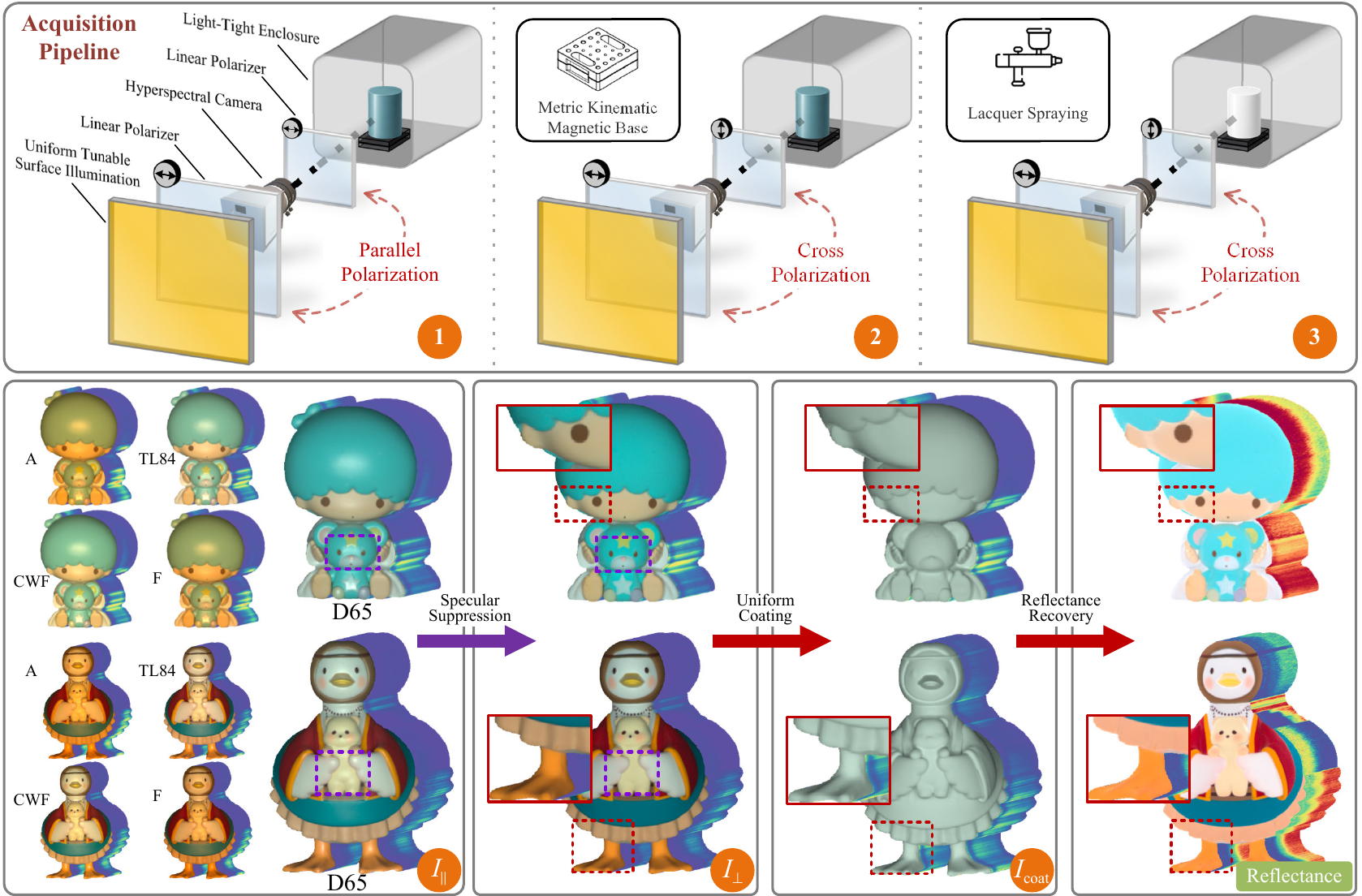}
  \caption{Overview of the CITE data acquisition pipeline and representative annotations. \textbf{Top:} Three sequential capture steps performed inside a light-tight enclosure: (1)~parallel polarization retaining both diffuse and specular reflections, (2)~cross polarization suppressing the specular component, and (3)~cross polarization after lacquer spraying for a spatially uniform reference reflectance. A metric kinematic magnetic base is used for precise repositioning after spraying; the depicted object is illustrative only. \textbf{Bottom:} Two representative objects from CITE. For each object, the left panel displays $\boldsymbol{I}_{\parallel}$ captured under five illumination conditions (A, TL84, CWF, F, D65). Taking D65 as an example, the acquisition sequence proceeds from $\boldsymbol{I}_{\parallel}$ through specular suppression to $\boldsymbol{I}_{\perp}$, uniform coating to $\boldsymbol{I}_{\mathrm{coat}}$, and reflectance recovery via Eq.~\eqref{eq:gt_derivation} to $\boldsymbol{R}$. \textcolor[RGB]{128, 12, 214}{Purple} boxes mark regions with notable specular reflections; \textcolor[RGB]{192, 0, 0}{red} boxes, shown with zoomed insets, highlight prominent shading variations.}
  \label{fig:DataSet}
\end{figure*}

\section{Dataset}
Several real-world HID datasets have been constructed for intrinsic scene analysis; however, existing public collections do not provide ground-truth annotations for specular-related photometric quantities, and some are further limited by the spectral fidelity of the acquisition device. Moreover, the variability arising from diverse illuminant spectra and spatially varying diffuse--specular mixtures encountered in real scenes is difficult to systematically span through direct acquisition alone, as each capture session is bound to a specific illuminant spectrum and the surface specularity of the imaged objects.

To address these gaps, we establish CITE, a real-world measurement set of DRM-related parameters for non-Lambertian objects. Building upon these high-fidelity measured quantities, we further develop a Physically-faithful Intrinsic Set Generator~(PISG) that enables controllable variation in illuminant spectra and diffuse--specular configurations. The PISG substantially enlarges the available data distribution beyond what physical acquisition can cover while preserving consistency with real-world physical distributions. The acquisition pipeline and the PISG are detailed in the following subsections.

\subsection{Data Acquisition}
\label{subsec:data_acquisition}
The CITE measurements were acquired using a Specim push-broom hyperspectral camera covering 80 spectral channels from 440 to 700\,nm. In total, 41 non-Lambertian objects with varying geometry and reflectance distributions were captured under five illumination conditions with distinct spectral power distributions, which increase the diversity of illuminant spectra and provide cross-condition observations for subsequent analysis. 

Following the cross-polarization protocol of Grosse~et~al.~\cite{grosse2009ground}, three pixel-aligned HSIs are acquired for each object under a fixed imaging geometry, as illustrated in Fig.~\ref{fig:DataSet}:
\begin{enumerate}
    \item \emph{Parallel-polarization image} 
     \( \boldsymbol{I}_{\parallel} \): the transmission axes of the two linear polarizers are aligned in parallel, so that the recorded HSI retains both diffuse and specular reflections.
    \item \emph{Cross-polarization image}
    \(\boldsymbol{I}_{\perp}\): the polarizers are adjusted to orthogonal orientations to suppress the specular reflection, producing a diffuse-only observation.
    \item \emph{Coated image}
    \(\boldsymbol{I}_{\mathrm{coat}}\): while maintaining the cross-polarized configuration, the object surface is uniformly coated with a thin layer of lacquer and re-imaged. The coated surface provides a spatially uniform reference reflectance spectrum, denoted by \(\boldsymbol{R}_{\mathrm{coat}}\).
\end{enumerate}

Dark-current subtraction and polarizer transmittance correction are performed during data processing. From these three captures, the DRM component annotations are derived as
\begin{equation}
\label{eq:gt_derivation}
\begin{aligned}
  \boldsymbol{R}
    &= \frac{\boldsymbol{I}_{\perp}\odot
            \boldsymbol{R}_{\mathrm{coat}}}
           {\boldsymbol{I}_{\mathrm{coat}}},\\[3pt]
  \boldsymbol{g}
    &= \left\langle
       \frac{\boldsymbol{I}_{\mathrm{coat}}}
            {\boldsymbol{R}_{\mathrm{coat}}\odot\boldsymbol{L}}
       \right\rangle_{\lambda},\\[3pt]
  \boldsymbol{k}
    &= \left\langle
       \frac{\boldsymbol{I}_{\parallel}
             - \boldsymbol{I}_{\perp}}
            {\boldsymbol{L}}
       \right\rangle_{\lambda}.
\end{aligned}
\end{equation}
where $\boldsymbol{L}$ denotes the measured illuminant spectrum, obtained by imaging a standard Lambertian reference sphere under the same illumination condition. In channels where $ \boldsymbol{I}_{\parallel} $ exhibits clipping distortion due to the sensor dynamic range limit, the specular term $\boldsymbol{I}_{\parallel}-\boldsymbol{I}_{\perp}$ is recovered by exploiting the measured illumination spectrum together with the remaining unsaturated spectral bands. Representative annotations are shown in Fig.~\ref{fig:DataSet}.

\subsection{Physically-faithful Intrinsic Set Generator}
\label{subsec:pisg}
Building upon the intrinsic set of reflectance and photometric components from CITE, the PISG generates non-Lambertian HSIs under controllable illuminant spectra and diffuse--specular configurations. Concretely, given the measured reflectance $\boldsymbol{R}$, shading factor $\boldsymbol{g}$, specular coefficient $\boldsymbol{k}$, a target illuminant spectrum $\boldsymbol{L}^{*}$, and two scalar control parameters $a$ and $b$, the PISG synthesizes a hyperspectral image as
\begin{equation}
\label{eq:pisg}
  \boldsymbol{I}
    = \Theta\!\bigl(
        \boldsymbol{g}^{*}\odot\boldsymbol{L}^{*}
        \odot\boldsymbol{R}
        \;+\;
        \boldsymbol{k}^{*}\odot\boldsymbol{L}^{*}
      \bigr),
\end{equation}
where
\begin{equation}
\label{eq:pisg_scaling}
  \boldsymbol{g}^{*}
    = \frac{a}
           {\lVert\boldsymbol{g}\odot\boldsymbol{L}^{*}
                  \odot\boldsymbol{R}
            \rVert_{\infty}}\;\boldsymbol{g}\,,
  \qquad
  \boldsymbol{k}^{*}
    = \frac{b}
           {\lVert\boldsymbol{k}\odot\boldsymbol{L}^{*}
            \rVert_{\infty}}\;\boldsymbol{k}\,.
\end{equation}

The normalization in Eq.~\eqref{eq:pisg_scaling} rescales the shading factor and specular coefficient so that $a$ and $b$ directly control the peak intensity contributions of the diffuse and specular terms, respectively, enabling intuitive adjustment of the diffuse--specular balance independently of the chosen illuminant spectrum. The operator $\Theta(\cdot)=\min(\cdot,\,1)$ applies element-wise clipping, reproducing the saturation effect that arises in real acquisitions when strong specular reflections exceed the sensor dynamic range. Because the reflectance is preserved and the associated DRM quantities are specified by construction, each synthesized HSI is naturally paired with the corresponding component-level annotations. The three control variables $\boldsymbol{L}^{*}$, $a$, and $b$ thus provide independent axes of variation in illuminant spectrum and diffuse--specular contributions, allowing systematic coverage of conditions that are difficult to exhaust through physical acquisition alone. Moreover, the use of real-world measured components ensures that the synthesized data preserves the material and geometric characteristics of actual objects.


\section{Experiments}
\label{sec:experiments}
This section evaluates the proposed DichroicFormer framework. We first describe the experimental setup and evaluation metrics, then compare with representative baselines through quantitative and qualitative analyses. Cross-dataset experiments on external real-world HSIs are further conducted to assess the generalization capability of the framework beyond the CITE acquisition conditions and to support the physical fidelity of the PISG-generated data. Finally, ablation studies isolate the contributions of each core module and key design choice.

\subsection{Experimental Setup}
\label{subsec:experimental_setup}
DichroicFormer is implemented in PyTorch and trained on a single NVIDIA RTX 4090D GPU using the LAMB optimizer with a batch size of~4 and input patches of size $256\times256$. On the CITE dataset, a group-wise split assigns 33~objects for training and 8~for testing, ensuring strict object-level separation. The model is trained for 150K iterations with a cosine-annealed learning rate initialized at $8\times10^{-4}$. From the 80 acquired channels (440--700\,nm), we discard the short-wavelength end below 450\,nm due to low signal-to-noise ratio and subsample 36 approximately uniformly spaced bands for training and evaluation. Training pairs are generated online by the PISG (Sec.~\ref{subsec:pisg}), where illuminant spectra are drawn from a pool of 30 representative spectral power distributions covering artificial and daylight sources, and the diffuse and specular scaling parameters $a$ and $b$ are sampled from $[0.65,\,0.85]$ and $[0.5,\,1.1]$, respectively, to cover representative diffuse--specular configurations observed in real acquisitions.

For quantitative evaluation, the PISG generates 240 test pairs from the held-out objects by combining five illuminant spectra with six uniformly sampled diffuse--specular configurations, spanning diverse illuminant and diffuse--specular contributions. Because existing public HID datasets are restricted to Lambertian or near-Lambertian settings, cross-dataset experiments instead adopt publicly available HSIs originally collected for spectral reconstruction, which offer non-Lambertian scenes from independent acquisition setups. As these datasets lack component-level ground truth, they serve exclusively for qualitative assessment and are adapted to the model input format accordingly.

We compare against four optimization-based intrinsic decomposition methods: Gu~et~al.~\cite{gu2013efficient}, Chen~et~al.~\cite{chen2017intrinsic}, Huang~et~al.~\cite{huang2018multispectral}, and Krebs~et~al.~\cite{krebs2020intrinsic}. Among these, Chen~et~al.\ and Huang~et~al.\ address only the diffuse photometric components and do not estimate specular quantities. To isolate decomposition performance from illuminant estimation errors, we provide the ground-truth illuminant spectrum to all baselines, whereas DichroicFormer operates without such information at inference time. We additionally report results from representative task-specific estimators for global illuminant estimation (Gray-Edge~\cite{khan2017illuminant}, White-Patch~\cite{land1971lightness}, ISNL~\cite{su2018illumination}, Shades-of-Grey~\cite{finlayson2004shades}) and specular separation (KL~\cite{gu2011specularity}, TI~\cite{tan2005separating}) to provide reference performance on individual subtasks.

\subsection{Evaluation Metrics}
Under the DRM, the component scales are mutually constrained: the illuminant spectrum, shading factor~$\boldsymbol{g}$, specular coefficient~$\boldsymbol{k}$, and reflectance cannot be rescaled independently without violating the image formation equation (see the Supplementary Material for a formal analysis). Standard per-component scale-invariant metrics, however, align each quantity in isolation, which not only conceals relative scale violations between~$\boldsymbol{g}$ and~$\boldsymbol{k}$, but may itself break the coupled scale relationship required by the image formation model. We therefore adopt a scale-coupled evaluation protocol. For reflectance, which admits no scale ambiguity under the joint constraint, we report absolute MSE and SSIM to assess decomposition fidelity without scale alignment. For the shading factor~$\boldsymbol{g}$ and specular coefficient~$\boldsymbol{k}$, we further introduce the Relative Scale Mean Squared Error (rs-MSE) defined under a shared positive scale factor:
\begin{small}
\begin{equation}
\label{eq:rsmse}
\begin{gathered}
\mathrm{rs\text{-}MSE}(\hat{\boldsymbol{g}})
  = \frac{1}{|\Omega|}\lVert \boldsymbol{g}_{\mathrm{gt}}
    - \beta\,\hat{\boldsymbol{g}} \rVert^{2},\quad
\mathrm{rs\text{-}MSE}(\hat{\boldsymbol{k}})
  = \frac{1}{|\Omega|}\lVert \boldsymbol{k}_{\mathrm{gt}}
    - \beta\,\hat{\boldsymbol{k}} \rVert^{2},\\[2pt]
\beta = \operatorname*{arg\,min}_{\beta > 0}
  \bigl(\lVert \boldsymbol{g}_{\mathrm{gt}}
    - \beta\,\hat{\boldsymbol{g}} \rVert^{2}
  + \lVert \boldsymbol{k}_{\mathrm{gt}}
    - \beta\,\hat{\boldsymbol{k}} \rVert^{2}\bigr),
\end{gathered}
\end{equation}
\end{small}
where $\Omega$ denotes the set of valid pixels and $|\Omega|$ its cardinality. The shared positive scalar~$\beta$ preserves only the global scale ambiguity common to both components, so that any relative scale discrepancy between them is directly penalized. The illuminant spectrum is evaluated with the Spectral Angle Mapper (SAM). To facilitate comparison with prior methods, we also report the per-component scale-invariant metrics si-MSE and si-SSIM. Detailed definitions and computational procedures for all metrics are provided in the Supplementary Material.

\subsection{Quantitative Evaluation}

\begin{table*}[!t]
\centering
\caption{Quantitative comparison of intrinsic decomposition methods on the CITE dataset under per-component scale-invariant metrics (si-MSE, si-SSIM) and scale-coupled metrics (MSE, SSIM, rs-MSE). \textcolor{red}{\textbf{Red}} and \textcolor{blue}{\underline{blue}} indicate the best and second-best results, respectively. N/A indicates that a metric is unavailable, either because the corresponding component is not estimated or because rs-MSE is defined under a shared scale factor jointly fitted to both $\boldsymbol{g}$ and $\boldsymbol{k}$; -- indicates that the metric does not apply, as all baselines receive the ground-truth illuminant spectrum rather than estimating it.}
\label{tab:HIID_quant}
\setlength{\tabcolsep}{2.5pt}
\renewcommand{\arraystretch}{1.35}
\resizebox{\textwidth}{!}{%
\begin{tabular}{l cc cc cc cc c}
\toprule
\multirow{2}{*}{\textbf{Method}} &
\multicolumn{4}{c}{\textbf{Reflectance}} &
\multicolumn{2}{c}{\textbf{Shading Factor}~$\boldsymbol{g}$} &
\multicolumn{2}{c}{\textbf{Specular Coeff.}~$\boldsymbol{k}$} &
\multicolumn{1}{c}{\textbf{Illuminant Spectrum}} \\
\cmidrule(lr){2-5}\cmidrule(lr){6-7}\cmidrule(lr){8-9}\cmidrule(lr){10-10}
& \textbf{MSE $\downarrow$}
& \textbf{si-MSE $\downarrow$}
& \textbf{SSIM $\uparrow$}
& \textbf{si-SSIM $\uparrow$}
& \textbf{rs-MSE $\downarrow$}
& \textbf{si-MSE $\downarrow$}
& \textbf{rs-MSE $\downarrow$}
& \textbf{si-MSE $\downarrow$}
& \textbf{SAM $\downarrow$} \\
\midrule
Chen~et~al.~\cite{chen2017intrinsic} &
10.1064 & 0.0440 & 0.1092 & 0.5487 &
N/A & 0.0257 & N/A & N/A & -- \\
Huang~et~al.~\cite{huang2018multispectral} &
0.3014 & 0.0484 &
\textcolor{blue}{\underline{0.4400}} & 0.5705 &
N/A & 0.0312 & N/A & N/A & -- \\
Gu~et~al.~\cite{gu2013efficient} &
\textcolor{blue}{\underline{0.1885}} & 0.0378 &
0.3972 & 0.7354 &
0.0287 & 0.0287 &
\textcolor{blue}{\underline{0.0170}} &
\textcolor{blue}{\underline{0.0156}} & -- \\
Krebs~et~al.~\cite{krebs2020intrinsic} &
0.3416 &
\textcolor{blue}{\underline{0.0323}} &
0.0182 &
\textcolor{blue}{\underline{0.7426}} &
\textcolor{blue}{\underline{0.0181}} &
\textcolor{blue}{\underline{0.0181}} &
0.0177 &
\textcolor{blue}{\underline{0.0038}} & -- \\
DichroicFormer &
\textcolor{red}{\textbf{0.0067}} &
\textcolor{red}{\textbf{0.0061}} &
\textcolor{red}{\textbf{0.8400}} &
\textcolor{red}{\textbf{0.8415}} &
\textcolor{red}{\textbf{0.0043}} &
\textcolor{red}{\textbf{0.0043}} &
\textcolor{red}{\textbf{0.0013}} &
\textcolor{red}{\textbf{0.0007}} &
0.0310 \\
\bottomrule
\end{tabular}%
}
\end{table*}

Tab.~\ref{tab:HIID_quant} summarizes the quantitative comparison of all intrinsic decomposition methods on the CITE dataset under both the per-component scale-invariant metrics (si-MSE, si-SSIM) and our scale-coupled metrics (MSE, SSIM, rs-MSE). Since all baselines are provided with the ground-truth illuminant spectrum, the SAM column is applicable only to DichroicFormer.

DichroicFormer achieves the best performance on every evaluated component under both metric families. Under the scale-invariant si-MSE for reflectance, DichroicFormer outperforms the strongest baseline by a factor of approximately five; this gap widens to over an order of magnitude under the absolute MSE ($0.007$ vs.\ $0.189$), indicating that a substantial portion of the baseline error is masked by independent per-component scale alignment. The advantage is also reflected in the perceptual metric: DichroicFormer's SSIM remains within $0.2\%$ of its si-SSIM, whereas Krebs~et~al.~\cite{krebs2020intrinsic}, which attains the second-best si-SSIM, suffers a drop from $0.743$ to $0.018$ when scale alignment is removed.

For the photometric components~$\boldsymbol{g}$ and~$\boldsymbol{k}$, the si-MSE results show that DichroicFormer recovers the scale-invariant spatial structures of both components more accurately than any baseline. Under the stricter rs-MSE, DichroicFormer still attains the lowest error on both components, with the shading and specular rs-MSE more than $4.2\times$ and $13.2\times$ lower than the respective second-best results. By comparison, Krebs~et~al.\ attains the second-best si-MSE for both components, yet their specular rs-MSE rises to roughly five times the corresponding si-MSE ($0.018$ vs.\ $0.004$), indicating that the relative scale between the two factors is poorly maintained.

Tabs.~\ref{tab:si-MSE_Spec} and~\ref{tab:SAM_Ill} further compare specular and illuminant estimation against representative task-specific estimators. DichroicFormer achieves the best specular si-MSE and the lowest illuminant SAM, outperforming the strongest competing methods by about $4.5\times$ and $26.7\%$, respectively, despite jointly recovering all components within a unified framework.

\begin{table}[!t]
\centering
\caption{Specular separation comparison with
task-specific estimators on the CITE dataset (si-MSE).}
\label{tab:si-MSE_Spec}
\setlength{\tabcolsep}{4.5pt}
\renewcommand{\arraystretch}{1.35}
\begin{tabular}{l cccc}
\toprule
& TI~\cite{tan2005separating}
& KL~\cite{gu2011specularity}
& ISNL~\cite{su2018illumination}
& Ours \\
\midrule
\textbf{si-MSE $\downarrow$} &
\textcolor{blue}{\underline{0.0030}} &
0.0114 &
0.0041 &
\textcolor{red}{\textbf{0.0007}} \\
\bottomrule
\end{tabular}
\end{table}

\begin{table}[!t]
\centering
\caption{Illuminant estimation comparison with
task-specific estimators on the CITE dataset (SAM).}
\label{tab:SAM_Ill}
\setlength{\tabcolsep}{2.5pt}
\renewcommand{\arraystretch}{1.35}
\begin{tabular}{l ccccc}
\toprule
& \makecell{White-Patch\\\cite{land1971lightness}}
& \makecell{Shades-of-Grey\\\cite{finlayson2004shades}}
& \makecell{Gray-Edge\\\cite{khan2017illuminant}}
& \makecell{ISNL\\\cite{su2018illumination}}
& Ours \\
\midrule
\textbf{SAM $\downarrow$} &
0.0805 &
0.0754 &
\textcolor{blue}{\underline{0.0423}} &
0.1183 &
\textcolor{red}{\textbf{0.0310}} \\
\bottomrule
\end{tabular}
\end{table}

\subsection{Qualitative Evaluation}
\subsubsection{\textbf{HIID}}

\begin{figure*}[!t]
  \centering
  \includegraphics[width=0.94\textwidth,height=.95\textheight,keepaspectratio]{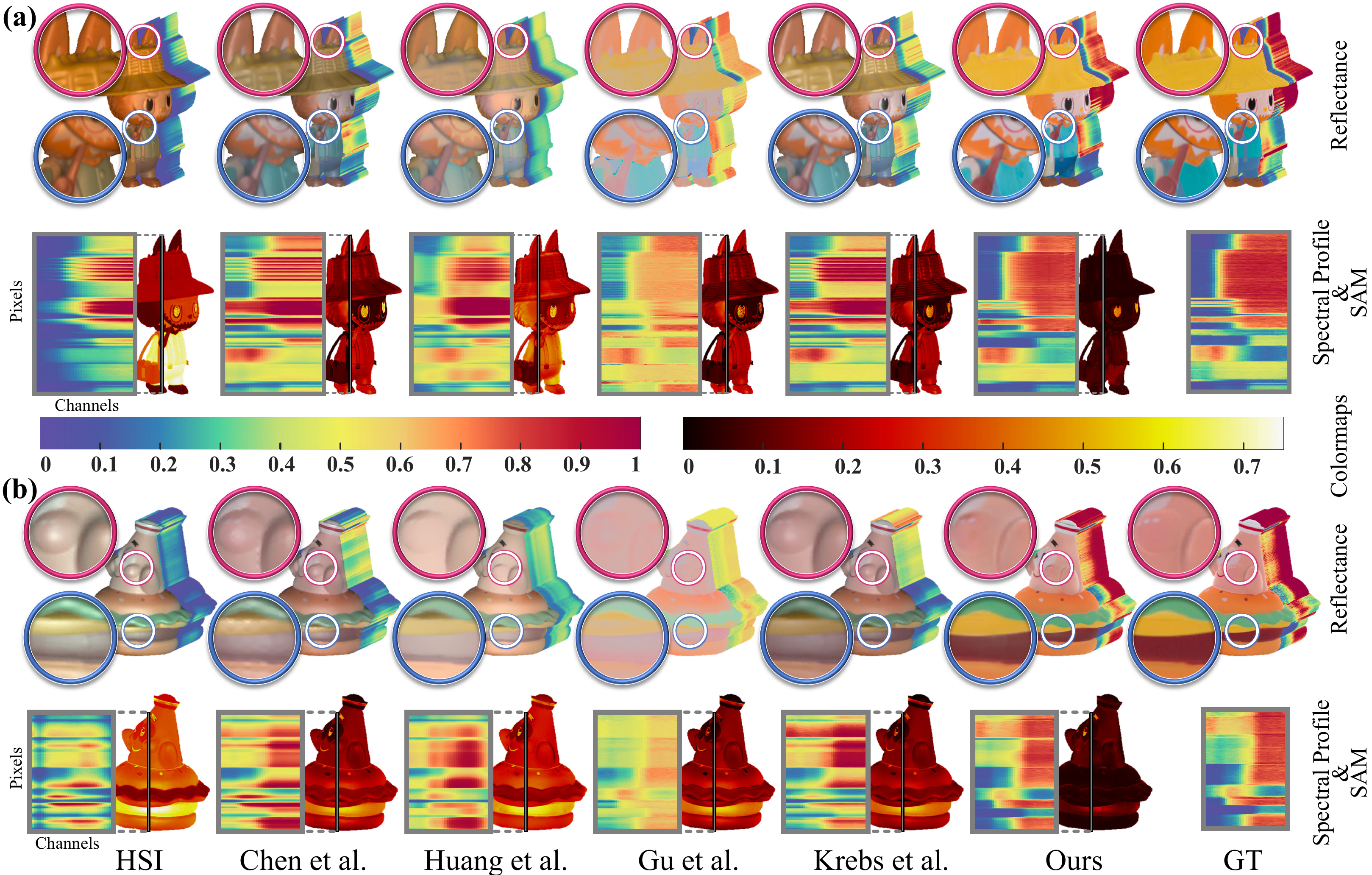}
  \caption{Qualitative comparison of reflectance recovery on two CITE test objects. Each object is shown with three complementary views: a three-dimensional pseudo-color rendering that unfolds the spectral dimension along the vertical axis with intensity-coded color, a normalized spectral profile map for the pixel column marked by the black line in the SAM map, and a per-pixel SAM error map with respect to the ground truth. Zoomed insets marked in \textcolor[RGB]{217,44,119}{pink} and \textcolor[RGB]{70,115,196}{blue} highlight representative regions of interest. The pseudo-color renderings and spectral profile maps share one colormap, whereas the SAM error maps use another, as indicated by the \textit{Colormaps} legend between the two objects. The ground-truth SAM map is omitted as it is identically zero.}
  \label{fig:HIID_Ref}
\end{figure*}

\paragraph{Reflectance}
Fig.~\ref{fig:HIID_Ref} compares reflectance recovery on two CITE test objects through pseudo-color renderings, spectral profile maps, and per-pixel SAM error maps. Under ideal decomposition, regions of the same material should exhibit consistent spectral profiles with stable inter-material discrimination, independent of local shading or specularity. Objects~(a) and~(b) are captured under warm- and cool-toned illumination, respectively, and both exhibit strong non-Lambertian effects, involving not only shading variations but also localized strong highlights and broadly distributed low-amplitude specularity. The input spectral profile maps further show that several of the brightest highlight regions remain at near-constant high levels across consecutive bands, consistent with clipping during acquisition.

In object~(a), Chen~et~al.~\cite{chen2017intrinsic} and Krebs~et~al.~\cite{krebs2020intrinsic} leave residual highlights in specularity-dominated regions such as the upright ear (pink inset) and the tobacco pipe (blue inset). Their SAM errors further spread over most of the surface, indicating incomplete removal of weak specular contamination. Huang~et~al.~\cite{huang2018multispectral} suppresses sparse strong highlights more effectively but exhibits a systematic spectral bias in the pseudo-color rendering with weakened inter-material contrast, yielding the most prominent SAM errors. All three methods show pronounced striping artifacts in the spectral profile map at pixels corresponding to the densely corrugated hat surface, indicating shading leakage into the reflectance. Gu~et~al.~\cite{gu2013efficient} mitigates such shading artifacts more effectively but at the cost of over-smoothing: the spectral profile map shows compressed band contrast, and intrinsic boundaries become blurred, most noticeably at the white cochlea texture.

Comparable observations arise in object~(b). In the pink inset of object~(b), Chen~et~al.\ and Krebs~et~al.\ show specularity-induced non-uniformity on the smooth cheek surface, while Huang~et~al.\ and Gu~et~al.\ flatten the fine reflectance texture around the eye socket. The blue inset covers the layered burger filling, where specular intensity varies continuously from strong to weak alongside pronounced shading. The compared methods either retain photometric residues or compress inter-layer spectral differences, producing unstable transitions in the spectral profile map.

DichroicFormer produces results closest to the ground truth, with sharper boundaries, cleaner local structures, and more consistent intra-material appearance across both objects. Its spectral profile map reproduces the band structure and inter-material contrast of the ground truth, and its SAM error remains the lowest and most spatially confined.

\begin{figure*}[!t]
  \centering
  \includegraphics[width=0.92\textwidth,height=.95\textheight,keepaspectratio]{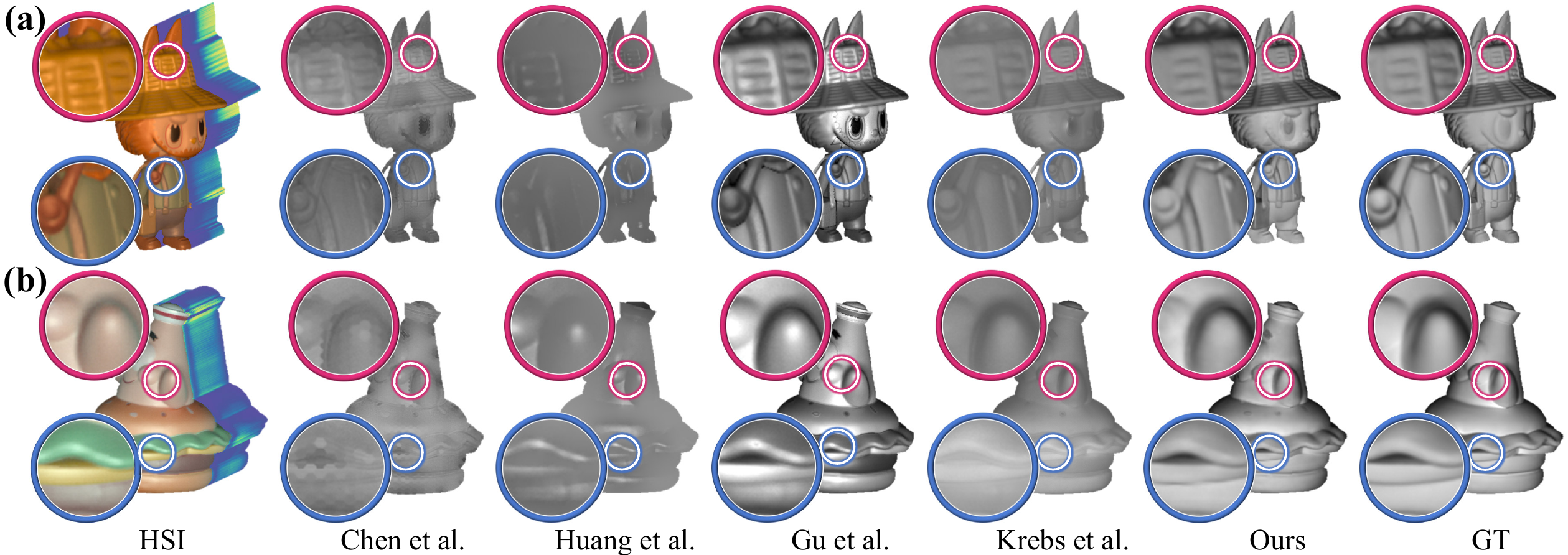}
  \caption{Shading factor estimates on two CITE test objects. Results are shown as grayscale visualizations. Zoomed insets marked in \textcolor[RGB]{217,44,119}{pink} and \textcolor[RGB]{70,115,196}{blue} highlight representative regions of interest.}
  \label{fig:HIID_g}
\end{figure*}

\begin{figure*}[!t]
  \centering
  \includegraphics[width=0.9\textwidth,height=.95\textheight,keepaspectratio]{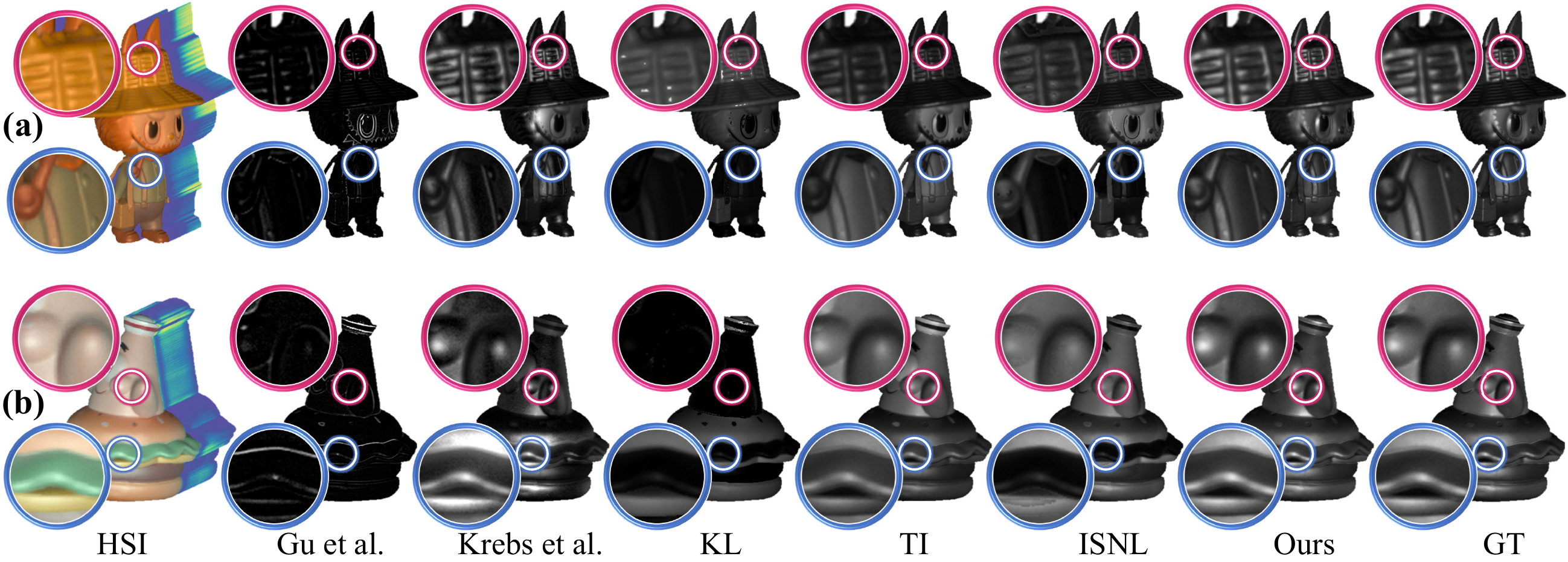}
  \caption{Specular coefficient estimates on two CITE test objects. This comparison includes both intrinsic decomposition baselines and task-specific specular separation methods. Each result is shown as a grayscale visualization. Zoomed insets marked in \textcolor[RGB]{217,44,119}{pink} and \textcolor[RGB]{70,115,196}{blue} highlight representative regions of interest.}
  \label{fig:HIID_k}
\end{figure*}

\paragraph{Shading}
Fig.~\ref{fig:HIID_g} further compares shading factor estimates of different methods. A desirable shading factor should preserve intensity variation induced by illumination and surface geometry while avoiding contamination from reflectance texture or specular highlights.

As the zoomed insets reveal, Chen~et~al.~\cite{chen2017intrinsic} produces physically implausible honeycomb-like artifacts on the corrugated hat of object~(a) (pink inset) and the burger filling of object~(b) (blue inset). Huang~et~al.~\cite{huang2018multispectral} yields globally over-darkened results with compressed tonal variation; the resulting flattened shading base makes the superimposed specular residues visually prominent, particularly on the clothing surface of object~(a) (blue inset). Gu~et~al.~\cite{gu2013efficient} better captures the shading variations induced by geometric structures, yet exhibits specular leakage, as evidenced by bright artifacts on the cheek in object~(b) (pink inset); additionally, the belly region of object~(a) shows intensity offsets indicative of reflectance leakage. Krebs~et~al.~\cite{krebs2020intrinsic} suppresses specular contamination more effectively, yet over-suppresses high-frequency geometric detail: the burger filling in object~(b) (blue inset), where stacked geometry produces pronounced shading transitions, is reduced to a nearly flat field. In contrast, DichroicFormer produces shading factors consistently closer to the ground truth. In geometrically detailed regions such as the corrugated hat of object~(a) and the cheek region of object~(b), the method preserves smooth geometry-related gradients while suppressing the artifacts and tonal compression observed in competing methods.

\paragraph{Specularity}
Fig.~\ref{fig:HIID_k} compares specular coefficient estimates, including both intrinsic decomposition baselines and task-specific specular separation methods. An accurate specular coefficient should capture both broadly distributed low-amplitude specularity and localized high-intensity highlights, with rapid yet smooth spatial decay away from highlight cores. 

Gu~et~al.~\cite{gu2013efficient} produces a largely under-responsive specular map, with only fragmented edge-aligned stripes rather than coherent specular structure. Krebs~et~al.~\cite{krebs2020intrinsic} overestimates the strong specular reflection regions, while the surrounding area where the ground truth shows smooth intensity decay instead exhibits granular fluctuations interspersed with abrupt dropouts to near-zero values, as evident on the cheek of object~(b) (pink inset). KL~\cite{gu2011specularity} produces an overly uniform map with localized intensity bias; the cheek of object~(b) (pink inset) is erroneously suppressed to nearly black, while the saturated highlight region on the hat of object~(a) (pink inset) contains anomalous white patches. TI~\cite{tan2005separating} localizes specular regions more plausibly, yet substantially compresses the amplitude range: weak specular responses are overestimated whereas strong peaks are underestimated, yielding a low-contrast map, as visible on the clothing of object~(a) (blue inset). ISNL~\cite{su2018illumination} underestimates peak specular intensity more severely than TI and introduces localized intensity offsets in the burger filling of object~(b) (blue inset), where artifacts also appear within the saturated highlight region. DichroicFormer shows the closest agreement with the ground truth. It faithfully recovers both the broad low-amplitude specularity on the clothing of object~(a) and the concentrated highlights on the raised burger filling surfaces of object~(b), whose post-peak spatial decay closely matches the ground truth.

\begin{figure*}[!t]
  \centering
  \includegraphics[width=0.9\textwidth,height=.95\textheight,keepaspectratio]{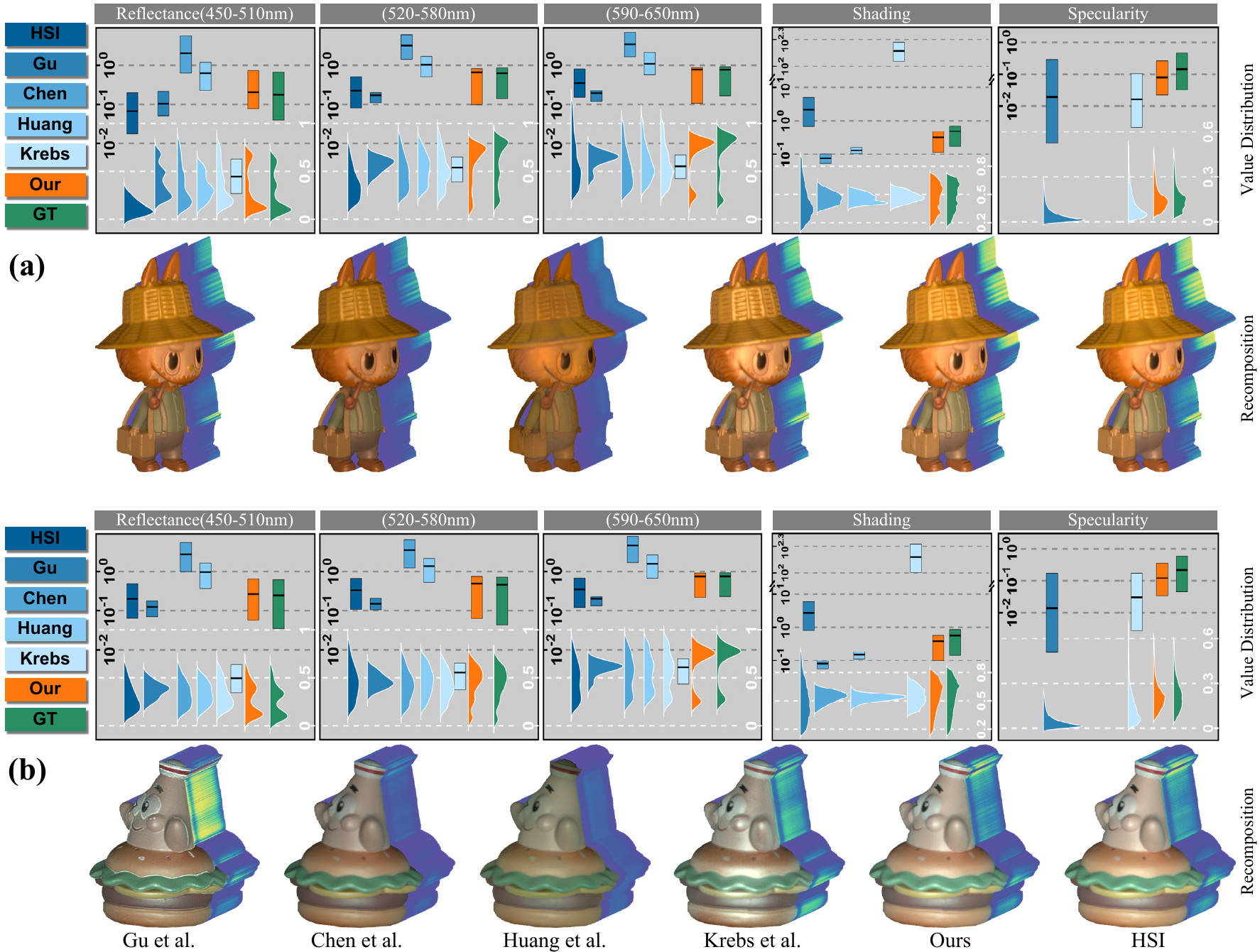}
  \caption{Component-scale analysis on CITE. For each of the two test objects, the statistical panels summarize the distributions of the decomposed components, including reflectance over three broad wavelength ranges (450--510 nm, 520--580 nm, and 590--650 nm), shading factor, and specular coefficient. Simplified boxplots indicate the median and the central 95\% range of nonzero pixels under the native output scale, read against the left logarithmic axis, whereas kernel density estimates are computed after per-component scale alignment to the ground truth, read against the right linear axis, and therefore reflect distributional shape. The input HSI is included in the reflectance-related panels as a reference. The recomposition panels visualize the images reconstructed from each method's independently aligned components according to the corresponding image formation model, using the same pseudo-color scheme as the reflectance visualizations. To better reveal inter-method differences, the display order of methods is adjusted from that used in other figures.}
  \label{fig:HIID_Scale}
\end{figure*}

\paragraph{Component-Scale Consistency}

Fig.~\ref{fig:HIID_Scale} examines scale fidelity from two complementary perspectives, namely per-component distribution statistics and object-level recomposition, both of which exhibit highly consistent trends across the two objects. In the statistical panels, a dual-axis layout pairs boxplots under the native output scale with kernel density estimates (KDEs) after per-component scale alignment, thereby separating native scale placement from distributional shape.

For reflectance, Chen~et~al.~\cite{chen2017intrinsic} is most strongly biased high, followed by Huang~et~al.~\cite{huang2018multispectral}, whereas Krebs~et~al.~\cite{krebs2020intrinsic} exhibits the largest downward shift and Gu~et~al.~\cite{gu2013efficient} displays severely compressed dynamic range, particularly in the middle- and long-wavelength bands. The KDEs reveal that these biases extend beyond simple scale mismatch: the ground-truth reflectance distributions exhibit broad support and distinct profiles, with peak structures that vary markedly across wavelength ranges. In contrast, Gu~et~al.\ collapses much of the distribution mass into a narrow interval, while the other baselines produce flatter, less distinctive curves that remain closer to the input HSI than to the ground truth, indicating limited disentanglement.

For the shading factor, Chen~et~al.\ and Huang~et~al.\ produce heavily compressed ranges biased low, whereas Gu~et~al.\ is biased high and Krebs~et~al.\ is two orders of magnitude above the ground truth. The KDEs mirror this tendency: Gu~et~al.\ exhibits an overly broad distribution while the remaining baselines are excessively concentrated, with Huang~et~al.\ showing the most severe compression, consistent with the flattened shading observed in Fig.~\ref{fig:HIID_g}. For the specular coefficient, Gu~et~al.\ and Krebs~et~al.\ both exhibit expanded dynamic range with downward intensity shift. The KDE of Gu~et~al.\ clusters near zero, corroborating the under-responsive map in Fig.~\ref{fig:HIID_k}. DichroicFormer closely tracks the ground truth in both scale magnitude and distributional shape across all components. As reflected jointly in the statistical panels and the quantitative scores of Krebs~et~al.\ (Tab.~\ref{tab:HIID_quant}), asynchronous, or even opposing, deviations of the shading and specular components from the ground truth can induce relative-scale mismatch that is obscured by per-component scale alignment; by contrast, rs-MSE exposes such inconsistency through scale competition under a shared scalar.

The recomposition panels further test consistency by recombining each method's independently scale-aligned components under the corresponding image formation model. Gu~et~al.\ underestimates the specular component and introduces edge artifacts, most notably along the upper boundaries of the layered burger filling in object~(b). Chen~et~al.\ yields recompositions that are globally too dark. Huang~et~al.\ exhibits pronounced spectral drift. Krebs~et~al.\ over-amplifies specular intensity. These recomposition deviations indicate that per-component scale-invariant evaluation may not reliably reflect decomposition quality, as independently rescaling each component can itself disrupt inter-component relative-scale relationships and introduce recomposition inconsistency. DichroicFormer yields recompositions visually closest to the input, further indicating that distributional fidelity is well preserved.

\begin{figure}[!t]
  \centering
  \includegraphics[width=0.88\columnwidth]{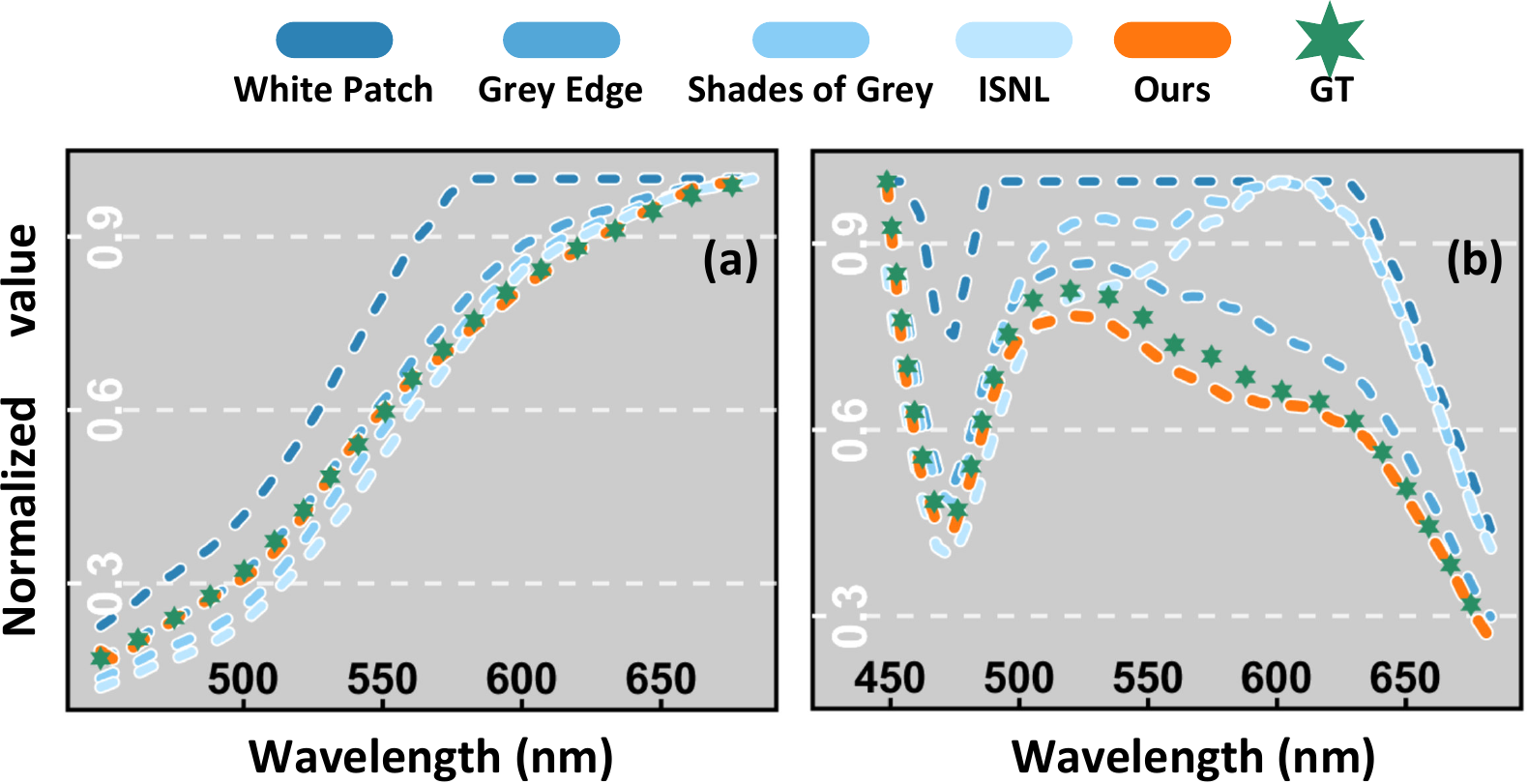} 
  \caption{Illuminant estimation on CITE. Normalized spectra estimated by DichroicFormer and four classical illuminant estimators are compared against the ground truth.}
  \label{fig:Illumination}
\end{figure}

\paragraph{Illuminant Spectrum}
Fig.~\ref{fig:Illumination} compares the normalized illuminant spectra estimated by DichroicFormer and four representative illuminant estimation methods on two test objects. DichroicFormer shows the closest agreement with the ground-truth spectral profile in both cases, despite recovering the illuminant jointly with all other components rather than as an isolated subtask. All baselines show less accurate estimates, with White-Patch exhibiting the largest deviation, as it relies on the maximum pixel response and is therefore most susceptible to sensor saturation.

\begin{figure*}[!t]
  \centering
  \includegraphics[width=0.92\textwidth]{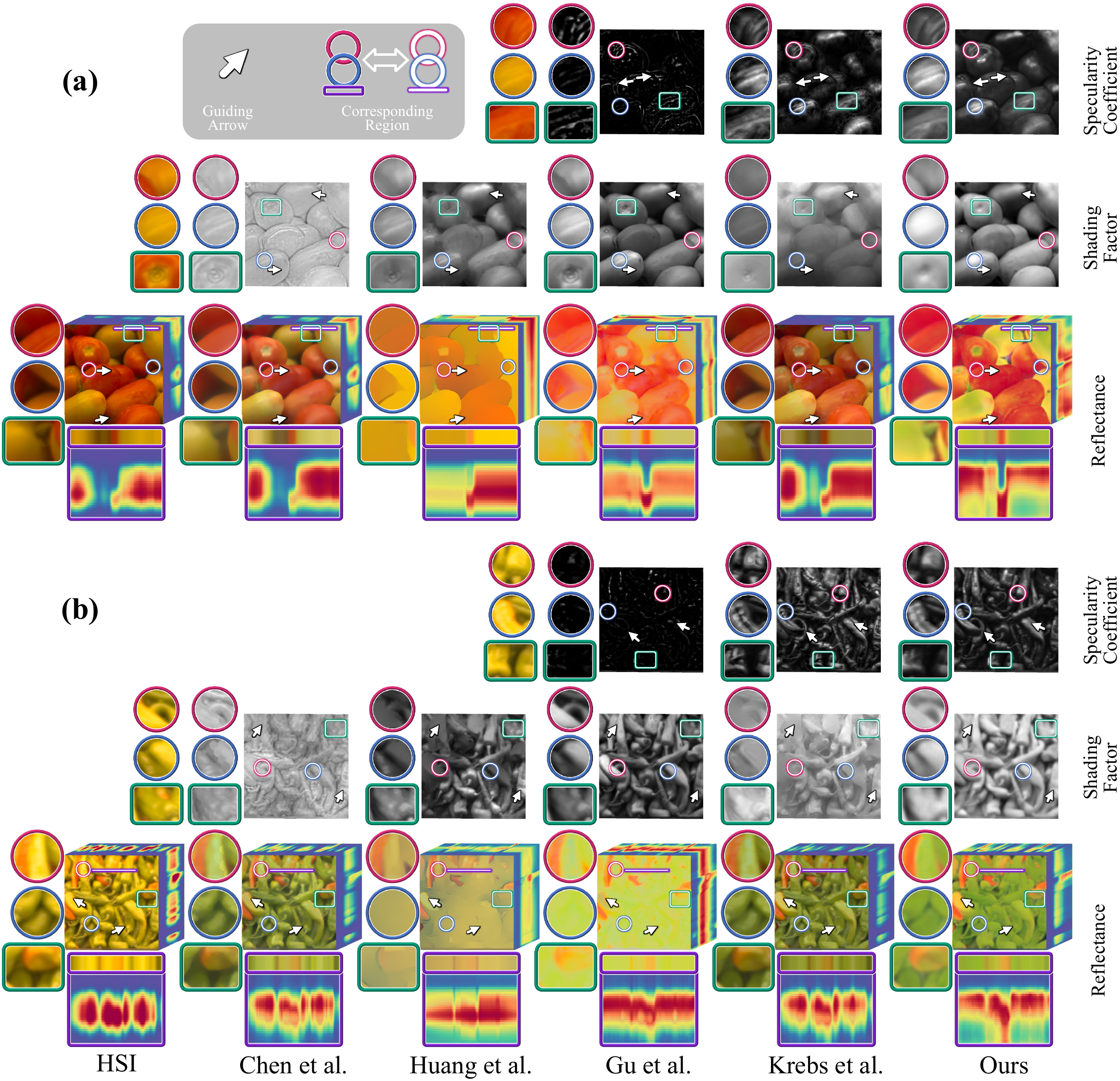}
  \caption{Qualitative intrinsic decomposition results on the KAUST dataset for two non-Lambertian scenes: (a)~tomatoes with fine-grained pigment variations and (b)~geometrically irregular peppers with overall stronger specular reflections. Each scene is organized into three rows corresponding to the specular coefficient, shading factor, and reflectance, respectively. In each reflectance column, the purple line marks a sampled line segment; the \textcolor[RGB]{140,36,225}{purple-bordered} panel below displays it as a magnified pseudo-color strip and an aligned spectral profile map. In the spectral profile map, the horizontal axis denotes spatial position along the sampled segment and the vertical axis denotes wavelength. Zoomed insets are marked in \textcolor[RGB]{204,40,108}{red}, \textcolor[RGB]{77,114,202}{blue}, and \textcolor[RGB]{0,156,118}{green}. The pseudo-color renderings and spectral profile maps adopt the same color mapping as Fig.~\ref{fig:HIID_Ref}. Arrows indicate additional regions of interest, including locations corresponding to inset regions in other component rows.}
  \label{fig:Kaust}
\end{figure*}

\subsubsection{\textbf{KAUST}}

The KAUST dataset was acquired with a portable Specim IQ push-broom camera ($512\times512$~pixels, 34 bands spanning 400--730 nm). The two representative examples in Fig.~\ref{fig:Kaust} present distinct non-Lambertian challenges: scene~(a) features tomatoes with fine-grained pigment variations across the surface, whereas scene~(b) contains geometrically irregular peppers with overall stronger specular reflections; the piling of objects in both scenes further enriches the shading variations.

\paragraph{Reflectance}
Huang~et~al.\ and Gu~et~al.\ produce reflectance estimates with over-smoothed spectral profiles that underestimate the reflectance variations induced by fine-grained pigment differences and introduce hues inconsistent with the expected material appearance (scene~(a) blue inset; scene~(b) green inset). This effect is particularly pronounced in scene~(b), where the spectral profile maps collapse into flattened horizontal bands and the central inter-material transition in the corresponding pseudo-color strips becomes blurred. Chen~et~al.\ and Krebs~et~al.\ preserve richer spectral diversity, yet residual specular reflections remain visible in concentrated highlight regions (scene~(a) green inset; scene~(b) blue inset). In scene~(a), the spectral profiles of both methods, together with the corresponding vertical dark streaks in the pseudo-color strip, further indicate shading leakage. The spectral profile maps of scene~(b) expose an additional saturation-related artifact shared by Chen~et~al., Gu~et~al., and Krebs~et~al.: in regions where the input spectra remain at near-constant high levels across consecutive bands, the recovered reflectance exhibits sustained suppression over multiple bands, forming localized anomalous patches. By comparison, DichroicFormer produces reflectance that faithfully captures the material properties: it maintains consistent reflectance among similarly ripe tomatoes (scene~(a) red inset), preserves clear spectral transitions and boundaries across pepper varieties (scene~(b) red inset), and remains stable in saturation-affected regions.

\paragraph{Shading Factor}

Chen~et~al.\ severely attenuates shading contrast, yielding an overly flattened decomposition that, in scene~(b), further breaks into patchy, mosaic-like discontinuities. As illustrated by the red inset of scene~(a), the estimates of Huang~et~al.\ and Gu~et~al.\ are visibly contaminated by specular leakage, exhibiting localized bright spots. Their estimates also exhibit excessively strong or weak shading transitions between adjacent objects (scene~(b) blue inset). Krebs~et~al.\ instead suppresses local variation excessively, leading to a smeared appearance over extended regions (scene~(b) red inset). Across both scenes, DichroicFormer yields shading fields that faithfully conform to the object geometry. This is most evident in the green inset of scene~(a), where only DichroicFormer effectively disentangles the concavity-induced shading fluctuation from stem-related reflectance variation and surrounding specular interference.

\paragraph{Specular Coefficient}
Gu~et~al.\ fails to capture most of the specular energy, yielding near-zero responses over large regions with only sparse bright patches remaining (scene~(a) blue inset; scene~(b) green inset). Krebs~et~al.\ captures the strong specular reflections more effectively but struggles with the soft specular spread, leaving regions beyond the highlight cores overly dark. As shown in the green inset of scene~(a), near-zero artifacts around the highlights further introduce intensity discontinuities that fragment the spatial distribution. By comparison, DichroicFormer produces specular estimates that closely follow the wrinkled tomato surfaces (scene~(a) red inset) and the elongated glossy ridges of the peppers (scene~(b) blue inset), yielding spatially coherent maps consistent with the surface normal variations.

Despite the differences in sensor, spectral sampling, and scene characteristics between KAUST and CITE, DichroicFormer consistently achieves the most physically coherent decomposition across all components. Together with the additional cross-dataset results in the Supplementary Material, these experiments further suggest that the proposed framework transfers reliably to external real-world HSIs, and also support the physical relevance of the PISG-generated training data beyond the CITE acquisition setup.

\begin{table}[!t]
\centering
\caption{Ablation of the global-stage backbone
organization. Metrics are computed at the global stage
only.}
\label{tab:ablation_backbone}
\setlength{\tabcolsep}{4pt}
\renewcommand{\arraystretch}{1.25}
\begin{tabularx}{\columnwidth}{
>{\centering\arraybackslash}m{0.22\columnwidth}
>{\centering\arraybackslash}m{0.14\columnwidth}
>{\centering\arraybackslash}m{0.26\columnwidth}
>{\centering\arraybackslash}m{0.26\columnwidth}
}
\toprule
\textbf{Structure} & \textbf{SR-CSR} &
\makecell{\textbf{Reflectance}\\\textbf{MSE}~$\downarrow$} &
\makecell{\textbf{Shading Factor}~$\boldsymbol{g}$\\\textbf{si-MSE}~$\downarrow$} \\
\midrule
Parallel    & --           & 0.0112 & 0.0074 \\
Branched    & --           & 0.0103 & 0.0068 \\
Serial      & --           & 0.0099 & 0.0056 \\
\midrule
Parallel    & $\checkmark$ & 0.0096 & 0.0073 \\
Branched    & $\checkmark$ & 0.0094 & 0.0056 \\
Serial      & $\checkmark$ &
  \textcolor{red}{\textbf{0.0085}} &
  \textcolor{red}{\textbf{0.0049}} \\
\bottomrule
\end{tabularx}
\end{table}


\begin{table*}[!t]
\centering
\caption{Ablation of the scale-specific modules (IDM,
SGAM) and the multi-scale progressive supervision within
the full dual-scale framework.}
\label{tab:ablation_components}
\renewcommand{\arraystretch}{1.25}
\begin{tabular}{
  l
  @{\hskip 28pt}
  c
  @{\hskip 28pt}
  c @{\hskip 16pt} c
  @{\hskip 28pt}
  c @{\hskip 16pt} c
}
\toprule
\multirow{2}{*}{\textbf{Variant}} &
\textbf{Reflectance} &
\multicolumn{2}{c@{\hskip 28pt}}{\textbf{Shading Factor}~$\boldsymbol{g}$} &
\multicolumn{2}{c}{\textbf{Specular Coeff.}~$\boldsymbol{k}$} \\
\cmidrule(l{-1pt}r{27pt}){2-2}
\cmidrule(l{-1pt}r{27pt}){3-4}
\cmidrule(l{-1pt}r{5pt}){5-6}
 & \textbf{MSE}~$\downarrow$
 & \textbf{rs-MSE}~$\downarrow$ & \textbf{si-MSE}~$\downarrow$
 & \textbf{rs-MSE}~$\downarrow$ & \textbf{si-MSE}~$\downarrow$ \\
\midrule
w/o IDM (no SR-CSR priors) & 0.00866 & 0.00546 & 0.00543 &
  0.00153 & 0.00100 \\
w/o SGAM guidance & 0.00819 & 0.00464 & 0.00462 &
  \textcolor{red}{\textbf{0.00118}} & 0.00075 \\
SGAM $\rightarrow$ CBAM~\cite{woo2018cbam} &
  0.00823 & 0.00478 & 0.00475 & 0.00137 & 0.00071 \\
w/o global-stage supervision & 0.00729 & 0.00511 & 0.00507 &
  0.00128 & 0.00075 \\
\midrule
DichroicFormer (full) &
  \textcolor{red}{\textbf{0.00670}} &
  \textcolor{red}{\textbf{0.00429}} &
  \textcolor{red}{\textbf{0.00425}} &
  0.00129 &
  \textcolor{red}{\textbf{0.00067}} \\
\bottomrule
\end{tabular}
\end{table*}

\subsection{Ablation Study}

We conduct ablation studies to isolate the contributions of four design choices: the global-stage backbone organization, the Invariant-Driven Module (IDM), the Specularity-Guided Attention Module (SGAM), and the multi-scale progressive supervision. Tab.~\ref{tab:ablation_backbone} evaluates backbone variants at the global stage, whereas Tab.~\ref{tab:ablation_components} ablates the remaining components on top of the selected backbone.

\subsubsection{\textbf{Backbone Architecture}} 
\label{sec:ablation_backbone} 

The proposed serial backbone cascades the diffuse estimate into the reflectance subnetwork, following the decomposition order of the inversion paradigm. To assess the effect of backbone organization, we compare against two alternative arrangements at the global stage:
\begin{itemize} 
\item \textbf{Parallel}: two independent subnetworks estimate the diffuse term and reflectance separately without intermediate coupling. 
\item \textbf{Branched}: a shared encoder feeds two decoder branches for the diffuse and reflectance estimates, respectively.
\end{itemize}

Serial and Parallel are parameter-matched, whereas Branched is lighter due to encoder sharing but uses the same number of transformer blocks.

As shown in Tab.~\ref{tab:ablation_backbone}, the serial design achieves the lowest reflectance MSE and shading si-MSE in all settings. The Parallel design keeps the two estimation paths entirely isolated, providing no mechanism for information flow between them. The Branched design partially alleviates this isolation through shared encoder features, but requires a single encoder to simultaneously serve quantities with different photometric dependencies, limiting component-specific encoding. By contrast, the serial cascade retains target-specific encoders while conditioning reflectance recovery on the diffuse estimate, thereby establishing cross-target interaction aligned with the decomposition order of the inversion paradigm and providing better-conditioned input for the subsequent reflectance estimation.

\subsubsection{\textbf{Invariant-Driven Module (IDM)}}
\label{sec:ablation_idm}

The IDM injects $\mathrm{SR}$ and $\mathrm{CSR}$ as paired edge-aware priors into the global-stage decoder, with each descriptor routed to the subnetwork whose target matches its level of photometric invariance (Sec.~\ref{sec:IDM}). We ablate them jointly: both descriptor inputs are removed while the module's layers and learnable parameters are retained, so that decoder feature propagation relies solely on implicit cues in the raw HSI.

Tabs.~\ref{tab:ablation_backbone} and~\ref{tab:ablation_components} show consistent degradation across all metrics upon removing the invariant priors. Notably, the dual-stage variant without priors (Tab.~\ref{tab:ablation_components}, \textit{w/o IDM}, reflectance MSE $= 0.00866$) is strictly worse than the single global stage equipped with SR-CSR (Tab.~\ref{tab:ablation_backbone}, serial with SR-CSR, MSE $= 0.0085$), despite having an additional refinement stage. This suggests that the quality of the global-stage decomposition directly constrains the effectiveness of the subsequent local refinement. By suppressing illuminant and specular contributions to different extents, $\mathrm{SR}$ and $\mathrm{CSR}$ provide target-aligned structural cues that are difficult to recover from implicit HSI features alone. The gains brought by SR-CSR are also more consistent under the proposed serial backbone, particularly on the global-stage shading si-MSE, suggesting that the explicitly ordered cascade provides a more suitable pathway for these invariant cues.

\subsubsection{\textbf{Specularity-Guided Attention Module (SGAM)}}

\label{sec:ablation_sgam}

The SGAM both modulates the backbone's encoder-to-decoder flow through specularity-aware features and supplies the final specular prediction through an independently supervised pathway. The following variants all retain the independent specular head (Sec.~\ref{sec:SGAM}) and assess the contribution of the specularity-aware guidance:

\begin{itemize}
\item \textbf{w/o SGAM guidance}: the specular head remains, but its features are no longer injected into the backbone.
\item \textbf{SGAM $\rightarrow$ CBAM}: the specularity-aware modulation is replaced with CBAM~\cite{woo2018cbam}, which performs channel-spatial reweighting derived from the backbone features themselves rather than from dedicated specular cues.
\end{itemize}

As shown in Tab.~\ref{tab:ablation_components}, the full model attains the best reflectance MSE, shading rs-MSE and si-MSE, and specular si-MSE. Compared with the single global stage (Tab.~\ref{tab:ablation_backbone}, serial with SR-CSR), the improvement is particularly pronounced on reflectance (MSE $0.0067$ vs.\ $0.0085$, $22\%$ reduction), indicating that the local stage substantially improves decomposition quality. Moreover, adding an unguided second stage yields only limited gain (e.g., reflectance MSE $0.0082$, $4\%$ reduction), suggesting that the contribution of the local stage lies less in merely adding an extra refinement stage than in the specularity-aware guidance it provides. Replacing this guidance with CBAM fails to reproduce the preceding gains. Although this variant employs a more elaborate attention architecture, it does not outperform the unguided variant on the main decomposition metrics. This further suggests that the gain depends more on the specular features used for guidance than on the attention mechanism design. The specular metrics remain broadly comparable across variants, with the full model attaining the lowest si-MSE and only small rs-MSE differences. This is consistent with all variants retaining the same independent specular head, whose gradients are decoupled from the backbone.

\subsubsection{\textbf{Multi-Scale Progressive Supervision}}

\label{sec:ablation_supervision}

We remove the global-stage loss terms while retaining supervision only at the local stage, leaving all architectural components intact. As shown in Tab.~\ref{tab:ablation_components}, removing global-stage supervision degrades reflectance MSE and both shading metrics. Without intermediate supervision, the propagation of intrinsic boundary cues becomes less effective, and the global stage may converge to degenerate representations that, while minimizing the final loss, lack meaningful photometric interpretation. As a result, the local stage is conditioned on less reliable global-stage outputs, weakening the benefits of SGAM-guided refinement. In this sense, the dual-scale scheme is effective not merely because it stacks two stages, but because the scale-specific modules and stage-wise supervision act together to enable a better-conditioned progressive decomposition.

\section{Conclusion}
Hyperspectral intrinsic decomposition under non-Lambertian conditions has long been constrained by simplified specular assumptions and the structural heterogeneity of the coupled reflectance and photometric components. To address these challenges, we propose DichroicFormer, a single-image framework built upon a unified inversion paradigm and a dual-scale decomposition scheme. The inversion paradigm reformulates the recovery of four heterogeneous DRM components around two spectral--spatial target variables, from which the associated photometric quantities can be derived in closed form, providing a compact basis for complete joint recovery without external auxiliary inputs. Building on this reformulation, the dual-scale scheme handles non-Lambertian effects of distinct spatial characteristics: the Invariant-Driven Module exploits the Spectral-Gradient Ratio and Cross Spectral-Gradient Ratio, which exhibit different levels of photometric invariance, as target-aligned edge priors for global intrinsic boundary preservation, while the Specularity-Guided Attention Module directs local refinement, with particular emphasis on specularity-dominated regions, through an independently supervised pathway that decouples the specular objective from the diffuse estimate. To support systematic benchmarking, we establish CITE, the first real-world HID dataset for non-Lambertian objects with component-level annotations; develop the Physically-faithful Intrinsic Set Generator for controllable synthesis under diverse illuminant spectra and diffuse--specular configurations; and introduce scale-coupled evaluation metrics that account for the coupled scale relationships inherent in the DRM. Extensive experiments and ablations on CITE and external real-world hyperspectral data confirm the effectiveness and cross-scene generalization of DichroicFormer. Promising directions include extending the framework to scenes with spatially varying illuminant spectra, generalizing across heterogeneous spectral sensors and acquisition setups, and integrating intrinsic decomposition with downstream hyperspectral analysis tasks.

\bibliographystyle{IEEEtran}
\bibliography{ref}  

\newpage 
\begin{IEEEbiographynophoto}{Hao Ye}
received the B.E. degree from the School of Computer Science and Engineering, Northeastern University, Liaoning, China, in 2022. 
He is currently pursuing the Ph.D. degree with the School of Electronic Science and Engineering, Nanjing University, Nanjing, China. His research interests include computational photography and computer vision.
\end{IEEEbiographynophoto}
\begin{IEEEbiographynophoto}{Zhan Shi}
received the B.E. degree from the School of Computer Science and Engineering, Northeastern University, Liaoning, China, in 2021. 
He is currently working toward the Ph.D. degree with the School of Electronic Science and Engineering, Nanjing University, Nanjing, China. His current research interests include computational photography, image processing, and hyperspectral imaging.
\end{IEEEbiographynophoto}
\begin{IEEEbiographynophoto}{Chenglong Huang}
received the B.S degree in communication engineering from Hohai University,China, in 2023. He is a graduate student from the School of Electronic Science and Engineering,Nanjing University.
\end{IEEEbiographynophoto}
\begin{IEEEbiographynophoto}{Tao Lv}
received the B.S degree from the College of Information Science and Engineering, Northeastern University, Liaoning, China, in 2021. He is currently working toward the Ph.D degree with the School of Electronic Science and Engineering, Nanjing University, Nanjing, China. His research interests include computational photography and computer vision, especially computational spectral imaging.
\end{IEEEbiographynophoto}
\begin{IEEEbiographynophoto}{Mingjie Ji}
is currently a Ph.D. candidate with the School of Electronic Science and Engineering, Nanjing University, Nanjing, China. His research interests focus on self-supervised learning and low-level computer vision, with a particular emphasis on video denoising.
\end{IEEEbiographynophoto}
\begin{IEEEbiographynophoto}{Qiu Shen} (Member, IEEE)
received the B.S. degree in electrical engineering and information science and the Ph.D. degree in signal and information processing from the University of Science and Technology of China, Hefei, China, in 2004 and 2009, respectively. From 2009 to 2016, she was with the Huawei 2012 Laboratory, Shenzhen, China, and Nanjing University of Aeronautics and Astronautics, Nanjing, China. 
She is currently a Faculty Member with the Electronic Science and Engineering School, Nanjing University, Nanjing. Her current research focuses on the next-generation video coding, collaborative video compression and analysis, and vision representation.
\end{IEEEbiographynophoto}
\begin{IEEEbiographynophoto}{Xun Cao} (Member, IEEE)
received the B.S. degree from Nanjing University, Nanjing, China, in 2006, and the Ph.D. degree from the Department of Automation, Tsinghua University, Beijing, China, in 2012. He held visiting positions at Philips Research, Aachen, Germany, in 2008, and Microsoft Research Asia, Beijing, from 2009 to 2010. He was a Visiting Scholar with The University of Texas at Austin, Austin, TX, USA, from 2010 to 2011. He is currently a Professor with the School of Electronic Science and Engineering, Nanjing University. His research interests include computational photography, image-based modeling and rendering, and virtual reality (VR)/augmented reality (AR) systems.
\end{IEEEbiographynophoto}

\end{document}